\documentclass[twoside,11pt]{article}

\usepackage{amsthm}
\usepackage{thmtools}
\usepackage{amsmath,amssymb}

\PassOptionsToPackage{hyphens}{url}
\usepackage[colorlinks=false,allbordercolors={1 1 1}]{hyperref}

\usepackage[capitalize]{cleveref}

\usepackage[square,comma,sort,numbers]{natbib}

\usepackage[utf8]{inputenc} 
\usepackage[T1]{fontenc} 
\usepackage{environ,comment}
\usepackage{amsfonts,bbm,bm,courier,nicefrac,microtype}
\usepackage{array,booktabs,tabularx,multirow,colortbl,hhline}
\usepackage{eqnarray}
\usepackage{enumitem}
\usepackage{xifthen}
\usepackage{thmtools}
\usepackage{xspace}
\usepackage{wrapfig}

\usepackage{thm-restate}

\usepackage{graphicx,xcolor,placeins}
\usepackage[font={footnotesize},labelfont={bf}]{caption}

\definecolor{good}{HTML}{BAFFCD}
\definecolor{bad}{HTML}{FFC8BA}

\newcolumntype{H}{>{\setbox0=\hbox\bgroup}c<{\egroup}@{}}
\newcolumntype{R}[1]{>{\raggedright\arraybackslash}p{#1}}

\usepackage{pifont}

\setitemize{leftmargin=1em,topsep=0.1em,parsep=0.5pt,}
\setlist[enumerate]{leftmargin=*, label= {\arabic*.}, itemsep=0.5em}

\usepackage{algorithm}
\usepackage{algpseudocode}

\algnewcommand{\alginput}[2]{\Statex{Input:~#1}\Comment{#2}}
\algnewcommand\algorithmicinput{\textbf{Input}}

\algnewcommand\algorithmicinitialize{\textbf{Initialize}}
\algnewcommand\algorithmicbigstep{\textbf{Step}}
\algnewcommand\INPUT{\item[\algorithmicinput]}
\algnewcommand\INITIALIZE{\item[\algorithmicinitialize]}
\algrenewcommand\algorithmiccomment[2][]{#1\hfill{\color{gray}\textit{\scriptsize{#2}}}}

\newcommand{\numberthis}{\addtocounter{equation}{1}\tag{\theequation}}

\usepackage{ext/jmlr2e_preprint}

\ifpdf
\hypersetup{
  pdftitle={Oversational Multiplicity},
  pdfauthor={E. George, D. Needell, B. Ustun}
}
\fi

\usepackage{etoolbox}

\usepackage{enumitem}

\newtheorem{theorem}{Theorem}
\newtheorem{remark}[theorem]{Remark}
\newtheorem{definition}[theorem]{Definition}
\newtheorem{lemma}[theorem]{Lemma}

\usepackage{lastpage}

\ShortHeadings{Observational Multiplicity}{George, Needell, and Ustun}
\firstpageno{1}

\DeclareMathOperator*{\argmin}{argmin}

\newcommand{\R}{\mathbb{R}}

\newcommand{\optpar}[2]{\ifthenelse{\isempty{#2}}{#1}{#1({#2})}}

\newcommand{\optsup}[2]{\ifthenelse{\isempty{#2}}{#1}{{#1}^{{#2}}}}

\newcommand{\optsub}[2]{\ifthenelse{\isempty{#2}}{#1}{{#1}_{#2}}}

\newcommand{\data}[1]{\optsup{\mathcal{D}}{#1}}

\newcommand{\X}{\mathcal{X}}

\newcommand{\clf}[1]{\optsub{f}{#1}}
\newcommand{\trueclf}[1]{\optsub{g}{#1}}

\newcommand{\eigmin}{\lambda_\mathrm{min}}
\newcommand{\eigmax}{\lambda_\mathrm{max}}

\newcommand{\mA}{\bm{A}}
\newcommand{\mB}{\bm{B}}
\newcommand{\mX}{\bm{X}}
\newcommand{\mH}{\bm{H}}

\newcommand{\pt}[1]{\bm{x}_{#1}}
\newcommand{\origlabel}[1]{y_{#1}}
\newcommand{\origlabelk}[2]{y^{(#2)}_{#1}}
\newcommand{\resamplabel}[1]{\hat{y}_{#1}}
\newcommand{\resamplabelk}[2]{\hat{y}^{(#2)}_{#1}}

\newcommand{\origdata}{\data{}}
\newcommand{\origdatak}[1]{\data{}^{(#1)}}
\newcommand{\resampdata}{\hat{\mathcal{D}}}
\newcommand{\resampdatak}[1]{\hat{\mathcal{D}}^{(#1)}}

\newcommand{\paramalg}{\mathcal{A}}

\newcommand{\paramvec}{\bm{\theta}}
\newcommand{\origparam}{\paramvec_*}

\newcommand{\resampparam}{\hat{\paramvec}_*}
\newcommand{\resampparamk}[1]{\hat{\paramvec}^{(#1)}_*}

\newcommand{\resamploss}[1]{\ell(#1; \resampdata)}

\newcommand{\origfunc}{\clf{\origparam}}

\newcommand{\resampfunck}[1]{\clf{\resampparamk{#1}}}

\newcommand{\Hess}{\mH}
\newcommand{\grad}{\bm{g}}
\newcommand{\invHess}{\Hess^{-1}}

\newcommand{\natureprob}[1]{p_{#1}^*}
\newcommand{\origprob}[1]{p_{#1}}
\newcommand{\origprobk}[2]{p_{#1}^{(#2)}}
\newcommand{\resampprob}[1]{\hat{p}_{#1}}
\newcommand{\resampprobk}[2]{\hat{p}^{(#2)}_{#1}}

\newcommand{\bernoullidist}{\textrm{Bern}}

\newcommand{\xmax}{X_{\max}}
\newcommand{\xmin}{X_{\min}}

\newcommand{\innerHinv}[2]{{\langle #1 , #2 \rangle}_{\invHess}}

\newcommand{\normHinv}[1]{{\| #1 \|}_{\invHess}}

\newcommand{\eventsmall}[1]{\mathcal{B}_{#1}}
\newcommand{\eventsmallcomp}[1]{\overline{\mathcal{B}_{#1}}}

\newcommand{\expec}{\mathbb{E}}
\newcommand{\prob}{\operatornamewithlimits{Pr}}
\newcommand{\sign}{\operatorname{sign}}
\newcommand{\transpose}{\intercal}
\newcommand{\Var}{\operatornamewithlimits{Var}}
\newcommand{\ind}{\bm{1}}

\newcommand{\defeq}{:=}

\crefname{hypothesis}{Hypothesis}{Hypotheses}
\crefname{fact}{Fact}{Facts}

\begin{document}

\title{Observational Multiplicity}

\author{\name Erin George \email e2george@ucsd.edu \\
       \addr Department of Mathematics\\
       University of California, San Diego\\
       La Jolla, CA  92093, USA
       \AND
       \name Deanna Needell \email deanna@math.ucla.edu \\
       \addr Department of Mathematics\\
       University of California, Los Angeles\\
       Los Angeles, CA 90095, USA
       \AND
       \name Berk Ustun \email berk@ucsd.edu \\
       \addr Department of Computer Science \& Halıcıoğlu Data Science Institute\\
       University of California, San Diego\\
       La Jolla, CA 92093, USA
       }

\editor{TBD}

\maketitle

\begin{abstract}
Many prediction tasks can admit multiple models that can perform almost equally well. This phenomenon can undermine interpretability and safety when competing models assign conflicting predictions to individuals. In this work, we study how arbitrariness can arise in probabilistic classification tasks as a result of an effect that we call \emph{observational multiplicity}. 
We discuss how this effect arises in a broad class of practical applications where we learn a classifier to predict probabilities $p_i \in [0,1]$ but are given a dataset of observations $y_i \in \{0,1\}$. 
We propose to evaluate the arbitrariness of individual probability predictions through the lens of \emph{regret}. 
We introduce a measure of regret for probabilistic classification tasks, which measures how the predictions of a model could change as a result of different training labels. 
We present a general-purpose method to estimate the regret in a probabilistic classification task.
We use our measure to show that regret is often higher for certain groups in the dataset and discuss potential applications of regret.
We demonstrate how estimating regret can be used to promote safety in real-world applications by abstention and data collection.
\end{abstract}

\vspace{0.25em}

\section{Introduction}
\label{Sec::Introduction}

Modern methods for machine learning fit models through \emph{empirical risk minimization} (ERM): given a training dataset with $n$ labeled instances $\origdata = \{(\pt{i}, \origlabel{i})\}_{i=1}^n$, we fit a model that minimizes the empirical risk. ERM is a optimization procedure: given a training dataset, running ERM multiple times should return models that achieve the same empirical risk. Given this behavior, it may be tempting to view the model that we recover as the ``best'' that we can do given the data that we have available. In practice, such claims would be wrong because there may exist multiple models that minimize the empirical risk. In effect, datasets can admit many ``competing models'' that perform equally well, i.e., model multiplicity ~\citep[][]{breiman2001statistical}. In such cases, these competing models can assign conflicting predictions to individual instances, i.e., \emph{predictive multiplicity} \citep[][]{marx2019predictive}.  When this happens, model predictions (and any downstream decisions made as a result) are dependent on the arbitrary choice of the model chosen from the set of competing models.

In applications like lending, recidivism prediction, and clinical prognosis---where we fit models that return probabilities---these claims would be wrong for a different reason: because the labels in the training datasets do not align with the outputs of the model. In such tasks, we would like to fit a model that outputs probabilities $\pt{i}\mapsto\prob(y_i=1 \;|\;\pt{i})$; however, our dataset contains labels $\origlabel{i}$ that represent realizations of these probabilities. In such cases, $\origdata$ represents a \emph{single random draw} of the labels in the training dataset $y_1,\ldots, y_n$. Had we been able to repeat this draw, we would have obtained an different training dataset $\origdata' = \{(\pt{i}, \origlabel{i}')\}_{i=1}^n$ which contains the same features but different observed labels $\origlabel{i}'$ for each point. 

\begin{figure}[t]
    \centering
    \resizebox{\linewidth}{!}{
    \begin{tabular}{lcr}
\multicolumn{1}{c}{\textsf{Original Dataset}} & 
\multicolumn{1}{c}{\textsf{Plausible Dataset 1}} & 
\multicolumn{1}{c}{\textsf{Plausible Dataset 2}}\\    
    \includegraphics[width=0.33\linewidth]{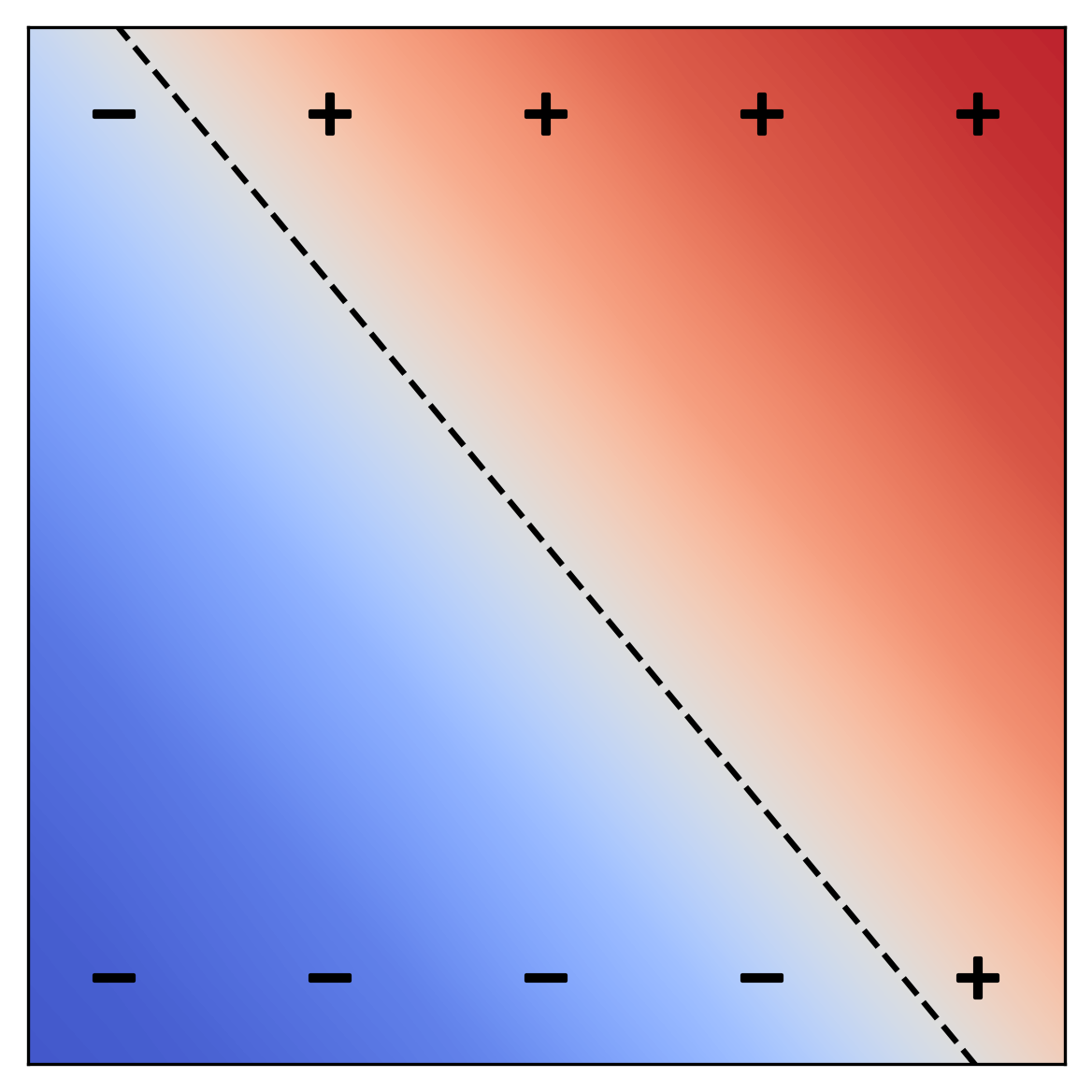}%
    &
    \includegraphics[width=0.33\linewidth]{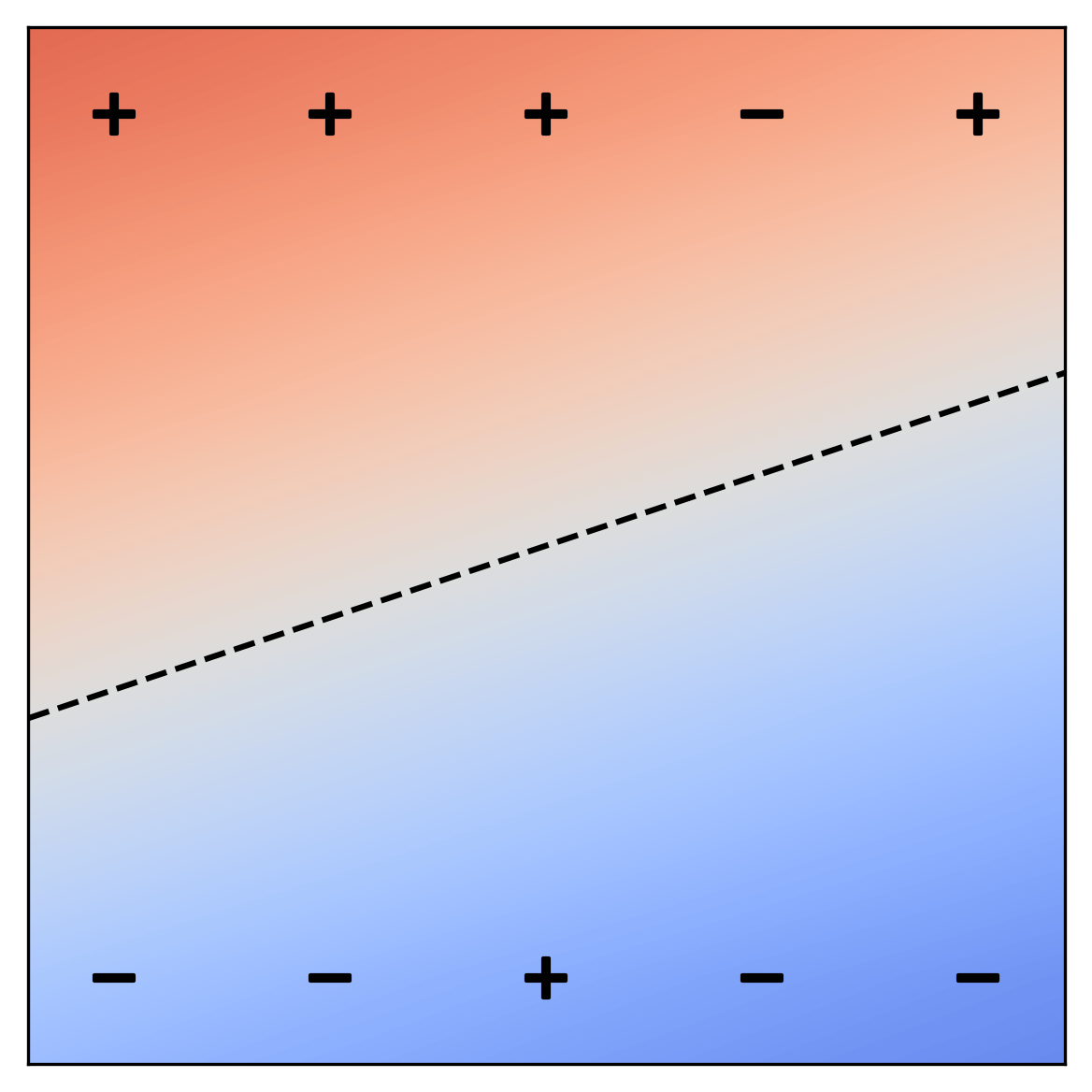}%
    &
    \includegraphics[width=0.33\linewidth]{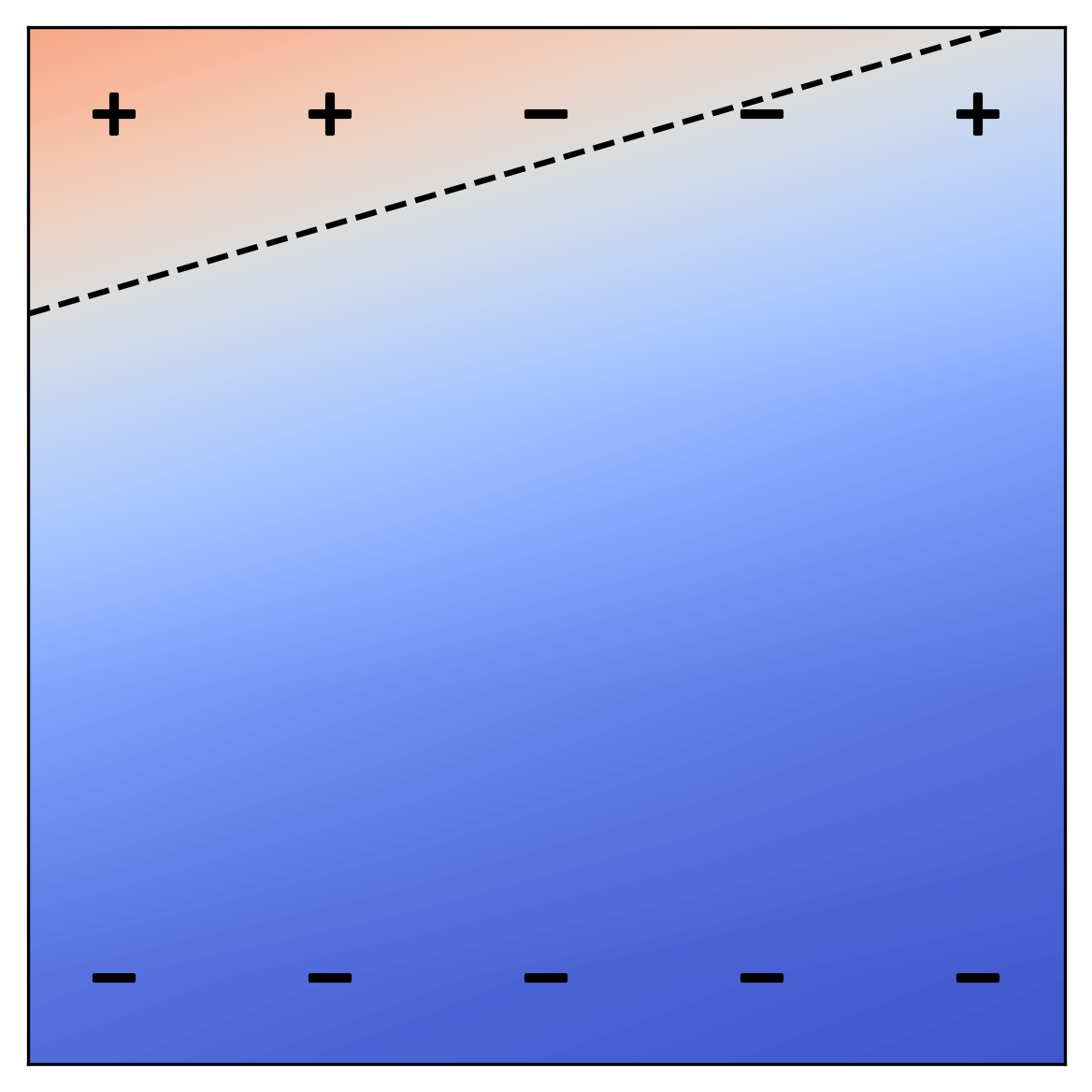} \\
    \multicolumn{1}{c}{Loss: 0.28, AUC: 1.0}& \multicolumn{1}{c}{Loss: 0.49, AUC: 0.84}& \multicolumn{1}{c}{Loss: 0.35, AUC: 0.90}
    \end{tabular}
    }
    \caption{Three different linear classifiers are obtained from different draws of the labels in a same dataset.  The points are plotted in the two-dimensional feature space.  The five points at the top and five points at the bottom all have an 80\% probability of lying in the positive and negative class, respectively.  There are four points that each lie on different sides of the decision boundary across the different draws.  The training cross-entropy loss and training AUC for each classifier is reported and varies substantially across draws.  No single dataset gives a full picture of the underlying pattern in the data.}
    \label{Fig::ObservationalMultiplicity}
\end{figure}

The datasets that we could produce through additional draws are equally plausible in a probabilistic classification task. It goes without saying that running ERM on $\data{}$ and $\data{}'$ would return different models $f$ and $f'$. 
In such settings, the models $f$ and $f'$ may perform almost equally well in terms of aggregate measures such as the logistic loss, AUC, and calibration, but this is not guaranteed.  Regardless of whether these quantitative measures agree, the procedure to obtain these models are identical.  If ERM to obtain one model was justified, it is equally justified to obtain the other.  Despite their shared validity, the models that we could obtain from alternative draws $f$ and $f'$ would assign different probability predictions to the same individual. 
In a standard data machine learning pipeline where the data is collected once, this randomness is unknown.  We are unable to see how different observations of the labels impact individuals (see e.g., \cref{Fig::ObservationalMultiplicity}).

\paragraph{Contributions} The main contributions of this work are as follows.

\begin{enumerate}

\item We define the term \emph{observational multiplicity} to refer to model multiplicity that arises from the data collection process.  We relate our definition to other forms of model multiplicity studied previously and demonstrate how it captures a different form of arbitrariness in model selection processes.

\item In analogy to online learning, we define the notion of \emph{regret} in the context of probabilistic classification: the unavoidable error that we incur due to the randomness in the observed labels.  We present a general-purpose method to estimate the regret in a probabilistic classification task, and show that regret is typically not evenly distributed throughout a population.

\item We demonstrate how our approach can be used to promote safety in real-world applications to aid in abstaining from arbitrary predictions or in collecting additional labels to reduce uncertainty.

\end{enumerate}

\paragraph{Related Work}

Our work is broadly related to a recent stream of work on multiplicity in machine learning~\citep[see e.g.,][]{breiman2001statistical,marx2019predictive,watson2023predictive,hsu2022rashomon,black2022model,damour2020underspecification,pmlr-v124-pawelczyk20a, meyer2023dataset,10.1145/3593013.3593998,ganesh2025systemizing,ICLR2025_94855277}. In particular, we describe a phenomenon called observational multiplicity that can lead to arbitrariness in individual predictions. Observational multiplicity is similar to predictive multiplicity as studied by \citet{marx2019predictive,watson2023predictive,hsu2022rashomon} in that both effects could be detected on validation datasets. In practice, however, they arise due to different mechanisms. 

Observational multiplicity is a specific variant of dataset multiplicity in that we can find competing models due to different plausible datasets~\citep[see][for other variants, such as sociologically biased datasets]{meyer2023dataset}. In contrast to other kinds of dataset multiplicity, it is particularly compelling because it affects a broad class of models (e.g., probabilistic classifiers) and because it occurs due to the inherent distributional assumptions we make when building probabilistic classifiers.

We propose to study observational multiplicity by introducing a measure of regret for probabilistic classification. Our measure captures how much individual probability estimate can change across plausible datasets. Our use of regret is motivated by that of \citet{nagaraj2025regretful}, who propose a measure of regret for classification task with label noise---i.e., where we learn a model to output hard labels $f(\pt{i}) \in \{-1,1\}$ from a dataset where the $\tilde{y}_i \in \{-1,1\}$ are corrupted through a noise model. In particular, we use the term ``regret'' reflect the fact that we are bound to incur some error in the individual predictions of a model given a fixed dataset. Our work develops a measure to capture these tasks when we build models that output probabilities (cf., models that output hard labels), and where datasets contain observations (cf., hard labels with label noise). 

A related concept in the study of machine learning is \emph{stability analysis}, which attempts to quantify how small changes in the training data set can propagate to the learned model~\citep{bousquet2002stability}.  Unstable predictors have large variance, which can be reduced using techniques such as bagging~\citep{buhlmann2002bagging}.  Stability has been studied in a variety of contexts, including support vector machines and linear regression~\citep{bousquet2002stability}, shallow neural networks~\citep{lei2022stability} graph convolutional neural networks~\citep{verma2019stability}, and transformers~\citep{li2023transformers}.

\paragraph{Organization}  We define and discuss observational multiplicity and regret in \cref{sec:obs_mult_discussion}.  \Cref{sec:theorems} contains the main theoretical results of the paper, regarding regret in logistic regression.  \Cref{Sec::Experiments} shows experiments on regret.  \Cref{sec:proof} contains the proofs of the theorems stated earlier in the paper.  \Cref{sec:conclusion} is the conclusion.

\section{Problem Statement}

\label{sec:obs_mult_discussion}

We consider a probabilistic classification task where we are given a dataset of $n$ labeled examples $\origdata = \{(\pt{i}, \origlabel{i})\}_{i=1}^{n}$. Each example consists of a feature vector $\pt{i} = [x_{i,1},\ldots,x_{i,d}] \in \X \subseteq \R^{d}$ and a label $\origlabel{i} \in \{-1, +1\}$. We assume that $\origlabel{i} = 1$ denotes an event of interest.  For example, $\origlabel{i} = 1$ may represent ``applicant $i$ will repay their loan'' or ``patient $i$ will develop cancer in the next five years''. We assume that the occurrence of this event is \emph{inherently stochastic}, i.e., that there is some randomness involved in the occurrence of the event of interest.%
\footnote{In view of philosophical concerns about whether true randomness can even be said to exist, we need only to assume that the labels are not completely determined by features and that this non-determinism cannot be resolved by adding a reasonable amount of additional features.  This nondeterminism may be a result of true randomness, but may also arise due to large amount of complexity in a system or exogenous factors.  However, referring to the labels as inherently stochastic provides useful intuitions.}

Our goal is to learn a \emph{probabilistic classifier} $\clf{}: \X \to [0,1]$, which takes as input a feature vector $\pt{i}$ and returns as output the predicted probability $\clf{}(\pt{i}) \defeq \text{Pr}(\origlabel{i} = +1 \;|\; \pt{i})$. 
We assume that each classifier is specified by a vector of \emph{parameters} $\paramvec \in \Theta$, and write $\clf{\paramvec}$ when the parameterization is relevant to the discussion. 

We fit the parameters of our model using some procedure $\paramalg$ that depends on the dataset $\origdata$ to obtain $\origparam = \paramalg(\origdata)$ and corresponding model $\origfunc \defeq \clf{\paramalg(\origdata)}$.  For example, we could use ERM with the loss function $\ell: [0,1] \times \{-1,1\}\to\R$ to obtain
\begin{align} 
\label{Opt::ERM}
\paramalg(\origdata) \in \argmin_{\paramvec \in \Theta} \sum_{i=1}^n \ell(\clf{\paramvec}(\pt{i}), \origlabel{i}).
\end{align}

\subsection{Observational Multiplicity}

Because the model is dependent on the dataset, both the model and its predictions will depend on the stochasticity inherent to the sampling procedure that yielded the initial training dataset.  Different datasets will yield different models. This is what we refer to as \emph{observational multiplicity}: the specific form of model multiplicity that arises downstream of the data sampling process.

Mathematically, suppose we have two separate draws of the training dataset: $\data{0}$ and $\data{1}$.  From these, we can obtain two distinct models $\clf{\paramalg(\data{0})}$ and $\clf{\paramalg(\data{1})}$. For this reason, we can view $\clf{}$ as a random variable, obtained as a function applied to the distribution of the sampled training set $\data{}$, and study the consistency of individual predictions.\footnote{This perspective has been used before to study consistency at the population level~\citep{zhao2006model, bach2010self}.}

This perspective is important because models are applied to individuals.  In many applications of machine learning, the subject of a classification task has a stake in the outcome of the classification problem.  The choice of one model over another will result in some people receiving different treatments (e.g., denied a loan that they otherwise might have received).  This is the standard predictive multiplicity issue.  As long as our model class is sufficiently expressive, this issue is unavoidable.  It is often trivial to create a new model which performs nearly identical to a base model except on one individual.

The way to ``resolve'' model multiplicity in practice is to select a model from the set of competing models (called the Rashomon set) according to some procedure, such as empirical risk minimization or stochastic gradient descent.  Crucially, these techniques used in practice do not fully resolve model multiplicity.  Even when the model selection procedure is deterministic once the dataset is given, the full pipeline of ``data collection and then fit model parameters'' is always a stochastic procedure.  As a process, it defines a distribution on the Rashomon set, and the final model is typically taken to be a single sample of this distribution.

It's important to know how concentrated the resulting distribution is.  Does this procedure truly resolve predictive multiplicity, or does it merely hide it?  This is more than just a mathematical curiosity: knowing the answer to this question can be used to inform a selective abstention procedure on top of the model when the predictions are found to be excessively arbitrary.

We remark that the consideration of varying datasets is essential to the full exploration of model multiplicity.  Typically, the notion of the Rashomon set is defined with respect to a specific reference dataset~\citep[see e.g.][]{marx2019predictive,hsu2022rashomon,watson2023predictive}.  We can think of a given dataset as a set of numerical constraints that define the Rashomon set.  When we change the dataset, we change the constraints, which results in a new Rashomon set.  However, the true quality of a model is determined by its performance on unseen data, not on a specified dataset.  A model is adopted because the selection process is thought to produce a model with good performance on unseen data.  Changing the dataset used to generate the model does not change the process itself, and so in principle the process produces an equally valid model even if the quantitative measures change.  Consider again \cref{Fig::ObservationalMultiplicity}.  The quantitative measure of the models vary drastically from one instance to the next, so a quantitative notion of the Rashomon set may exclude of the models from considerations.  However, all of these models were obtained from the same procedure on different datasets, so all of them were possible models to be chosen.

The distinction between observational multiplicity and previously studied forms of model multiplicity can be described as follows.  Past studies on model multiplicity have been concerned with quantitatively similar models without regard to the selection process, whereas in observational multiplicity we consider models with similar selection processes without regard to quantitative performance.  In previously considered settings, arbitrariness arises primarily as a result of the decision of a practitioner to select model in accordance to one procedure over another.  In our setting, the arbitrariness is a result of randomness in the circumstances outside of the practitioner's control.

As described, observational multiplicity is a phenomenon applicable to more than just probabilistic classification.  Any source of randomness (e.g., random sampling) in the data collection procedure will yield observational multiplicity.  However, we primarily concern ourselves with probabilistic classification in this paper because in this context there is a particularly significant and unavoidable source of stochasticity: the observed labels present in the initial training dataset.  We explore the consequences of this in the following subsection.

\subsection{On Regret}

In a probabilistic classification task we would like a model that outputs probabilities, but we can only train it from these observed labels, which are events that are a single draw of the underlying stochasticity.   From these labels we obtain our initial model $\clf{0} \defeq \clf{\paramalg{\data{0}}}$.  Suppose we were given the ``underlying model'' $\trueclf{}: \X \to [0,1]$: a probabilistic classification model that outputs the \emph{ground truth probability} $\natureprob{i} \defeq \trueclf{}(\pt{i})$ for each point $\pt{i} \in \X$.  Given the underlying model, we can ``redraw'' the labels by sampling
\begin{align}
    \origlabelk{i}{1} \sim \textrm{Bern}(\trueclf{}(\pt{i})).
\end{align}
We can then use this new set of observed labels to construct new dataset $\origdatak{1}= \{(\pt{i}, \origlabelk{i}{1})\}_{i=1}^{n}$ and obtain a new classifier $\clf{1} \defeq \clf{\paramalg(\data{1})}$.  

It is clear that $\clf{1}$ is a function of the random variables $\origlabelk{i}{k}$ and therefore so are the predicted probabilities $\origprobk{i}{1} \defeq \clf{1}(\pt{i})$.  This new draw comes from the same underlying distribution, so we would hope that our predicted probabilities from our model do not differ from our original predicted probability $\origprobk{i}{0} \defeq \clf{0}(\pt{i})$.  Of course, complete independence is too much to ask for: if the label is never a fully deterministic function of the features, then any label vector has a non-zero probability of being realized.

We will measure this uncertainity with a quantity we call the \emph{regret} we incur by training on a \emph{single draw of the labels}.  Regret is a measure that allows us to analyze a special case of observational multiplicity.
\begin{definition}
The \emph{regret} of a data point $\pt{i}$ in a model selection procedure $\paramalg(\cdot)$ on a dataset $\origdata$ for a probabilistic classification model $f_\theta(\cdot)$ is the variance of the random variable
\[f_{\paramalg(\origdata')}(\pt{i}),\]
where $\origdata'$ is a random dataset obtained by resampling the labels from $\origdata$ according to the true conditional probabilities of the labels given the data points.
\end{definition}

We use the term ``regret'' because, in analogy to online learning, it reflects unavoidable risk we incur because as a result of randomness---i.e., aleatoric uncertainty. In online learning, the aleatoric uncertainty arises because we cannot predict a random draw that will take place in the future. In our setting, the aleatoric uncertainty arises as a result of a random draw that took place in the past.  As in the online learning setting, we should not expect to ever achieve zero regret in a probabilistic classification task. We are bound to experience some regret because a single draw is insufficient to infer the probability distribution of a single label.  

There are two insights to be gained from this analysis.

\begin{remark}
Regret is unavoidable.
\end{remark}
The setting of probabilistic classification is exactly the same without access to the underlying model, except we only have access to a single draw and are not able to compute any measure of regret for it.  The regret still exists, so having a way to estimate the distribution of possible regrets is informative.

\begin{remark}
The burden of regret is not shared equally among the data points.
\end{remark}
When the probability given by the underlying model $\natureprob{i}$ is close to $0$ or $1$, there is less ``noise'' in the draw and the assigned probability will likely change less over the different samples compared to what happens when $\natureprob{i}=1/2$.  Additionally, if there is a cluster of data points with similar features and similar probabilities given by the underlying model, the empirical probability of $y = 1$ for this cluster will tend to concentrate around the mutually shared probability, and so the assigned probabilities will likely change less than for isolated points.  We will see later that the range of values regret takes over the dataset is often quite large.

\subsubsection{Estimating Regret}
\label{Sec::Estimation}

In an ideal world, to measure regret (or otherwise quantify observational multiplicity), we would collect fresh samples of the data and retrain the model multiple times.  Ordinarily, the process of collecting fresh data is merely impractical.  For regret, however, it often impossible.  In tasks where we can collect more data, we are often only able to sample the labels of \emph{new} individuals rather than from individual in our dataset---e.g., we cannot offer a new loan to someone previously in a loan-default dataset).  Under the belief that there was randomness involved in the observed label, this is a significant problem.  There is no way to truly get around the issue of the sample size being one.  We are forced to estimate regret using a counterfactual procedure.

\paragraph{Procedure} A general procedure to estimate regret is to obtain some approximation of the ``underlying model'' from the previous section and use this as a proxy to resample the labels ourselves.  This approximation, of course, is exactly the model we were attempting to learn in the first place.  We present this procedure in \cref{alg:regret}.  This algorithm is a form of parametric bootstrapping, and the general technique is sometimes called ``model-based resampling''~\citep[see e.g. Algorithm 6.1 in][]{bootstrapmethods}.
\begin{algorithm}[H]
\caption{Estimation Procedure}\label{alg:regret}
\begin{algorithmic}\small
\Require{$\origdata = \{(\pt{i}, \origlabel{i})\}$}\Comment{Training Dataset}
\State $\origparam \gets \paramalg(\origdata)$ \Comment{Fit initial model}
\State $\origprob{i} \gets \origfunc(\pt{i})$
\For{$k$}
\State $\resamplabelk{i}{k} \sim \bernoullidist(\origprob{i})$\Comment{Resample labels}
\State $\resampdatak{k} \gets \{(\pt{i}, \resamplabelk{i}{k})\}$
\State $\resampparamk{k} \gets \paramalg(\resampdatak{k})$ \Comment{Fit new model}
\State $\resampprobk{i}{k} \gets \resampfunck{k}(\pt{i})$\
\EndFor
\Ensure{$R_i \gets \Var(\{\resampprobk{i}{k}\}_k)$}\Comment{Estimated regret}
\end{algorithmic}
\end{algorithm}

Estimating regret is extremely difficult.  \Cref{alg:regret} does not produce an unbiased estimator of regret in general.\footnote{As a simple example, if we run \cref{alg:regret} on a non-probabilistic classifier for the problem, all labels will be deterministic and predicted regret will be $0$ for all individuals.} However, without any model assumptions, it difficult to quantify how sensitive a machine learning model is to its input data without sampling new input data \citep[see][for the case of stability analysis]{kim2023blackbox}.  In the case of regret, collecting new data is impossible.  Simulating new data is very difficult, as to do so we require knowledge of the ground truth probabilities $\natureprob{i}$, which is exactly the problem we are employing the probabilistic classifier to solve.  In absence of theoretical guarantees for the algorithm, we show experiments in \cref{sec:validation} that demonstrate how well \cref{alg:regret} works in practice.

\Cref{alg:regret} assumes that the dataset obtained by the resampling process is similarly plausible to the original dataset.  This assumption may not be true in practice, because the relationship between the features and the label may not be accurately represented by the model.  However, even in this situation the results of this resampling process are still meaningful.  This is because of when we deploy a model, we are simultaneously declaring some degree of confidence in its accuracy.  The resampled labels are, by construction, a plausible set of labels according to the model.  Therefore, the uncertainty measured by this procedure is directly implied by the model we are seeking to deploy.


\section{Theoretical Results}\label{sec:theorems}

With regret defined, we now are interested in determining properties of the measure.  We will use logistic regression as an example case, because for this model we can show a bound for the expected value of the regret estimator, which is given by the following theorem.
\begin{restatable}{theorem}{thmlogregprob}\label{thm:var_bd}
Suppose $n \geq 4$.  Consider a logistic regression model on the dataset $\resampdata = \{(\pt{i},\resamplabel{i})\}_{i=1}^n$ where the features $\pt{i}\in\R^d$ are fixed while the labels are sampled $\resamplabel{i}\sim\bernoullidist(\origprob{i})$ with $\origprob{i}$ given by an initial logistic regression model with parameters $\origparam$.  Define
\[Q_i \defeq \origprob{i}^2(1-\origprob{i})^2 \pt{i}^\transpose \mH^{-1}  \pt{i},\]
where
\[\mH \defeq \sum_j \origprob{j}(1-\origprob{j})\pt{j}\pt{j}^\transpose.\]
Assume that $\Hess$ is full rank and let
\[\epsilon \defeq 104\cdot\frac{d\xmax\left(\log(n\xmax/\xmin) + \xmax\|\origparam\|_2\right)}{\sqrt{\eigmin}}.\]
If $\epsilon < 1$, then the output probability $\resampprob{i}$ of $\pt{i}$ from the logistic regression model is a random variable satisfying
\begin{align*}
    |\Var(\resampprob{i}) - Q_i| \leq \epsilon Q_i.
\end{align*}
The terms $\xmax$ and $\xmin$ refer to the maximum and minimum values of $\|\pt{i}\|$ over all indices $i$, while $\eigmin$ is the minimum eigenvalue of the $d \times d$ matrix $\Hess$.
\end{restatable}
The proof of this theorem is in \cref{sec:proof} and follows from results shown by \citet{bach2010self}.  This section also contains further details on how the new logistic regression model is generated; however, we comment here that the method we consider is standard.

The quantity $Q_i$ is known in the statistics literature to be an estimate for the asymptotic variance of $\resampprob{i}$.  This arises from the delta method applied to the variance of the logits, which can estimated using the empirical information matrix $\mH$ \citep[see][Section 5.5.3 for variance of logits, Section 14.1 for delta method]{agrestibook}.  In our context, \cref{thm:var_bd} is a non-asymptotic bound for how close this approximation is.

If we sample the original data from a sufficiently nice distribution on $\R^d$, then many of the terms in the expression for $\epsilon$ from \cref{thm:var_bd} can be bounded by constants with high probability.  The following corollary describing the limiting behavior of $\epsilon$ is an example of this.
\begin{restatable}{corollary}{corofixeddist}\label{cor:fixed-dist}
Using the notation of \cref{thm:var_bd}, suppose the parameters $\origparam$ for the logistic regression model used to resample the labels are obtained by fitting the parameters on an initial dataset $\mathcal{D} = \{(\pt{i},\origlabel{i})\}_{i=1}^n$, where each row of the dataset is drawn independently from a fixed probability distribution $\mathcal{P}$ with compact support in $\R^d\setminus\{0\}$.  Suppose further that when $(\pt{},\origlabel{})$ is sampled from $\mathcal{P}$, the marginal distribution of $\pt{i}$ has full rank covariance matrix and there is a compact subset $K \subset (0,1)$ such that $\prob(\origlabel{}=1 \;|\; \pt{}) \in K$ for all $\pt{}$ when $(\pt{},\origlabel{})$.  

There exists constants $C_1(\mathcal{P})$ and $C_2(\mathcal{P})$ such that if $n \geq C_1(\mathcal{P})(1+\log(1/\delta))$, then with probability at least $1-\delta$ over the random sample $\mathcal{D}$ we have that
\[\epsilon \leq C_2(\mathcal{P}) \cdot \frac{\log n}{\sqrt{n}},\]
The constants $C_1(\mathcal{P})$ and $C_2(\mathcal{P})$ depend on $\mathcal{P}$ but not $n$.
\end{restatable}
A full proof of this corollary is in
\cref{app:cor-proof}.  This setting results in all eigenvalues of $\mH$ scaling proportionally to $n$, so $Q_i$ will scale as $\Theta(1/n)$.  Similar results to \cref{cor:fixed-dist} can be shown under weaker conditions for $\mathcal{P}$.

\Cref{thm:var_bd,cor:fixed-dist} apply both to estimated regret determined according to \cref{alg:regret} and the true regret we would obtain in a well-specified setting (in the case of true regret, $\origprob{i}$ and $\natureprob{i}$ take the role of $\resampprob{i}$ and $\origprob{i}$ in the theorem statement, respectively).  This theorem shows that regret is not distributed evenly across the points.  The regret for $\pt{i}$ is larger when $\origprob{i}$ is closer to $1/2$ or when $\pt{i}$ lies near an eigenspace of $\mH$ with small eigenvalue.

\paragraph{Example} Suppose the dataset consists of two groups of individuals that lie in orthogonal one-dimensional subspaces.  Group 1 contains $n_1$ individuals that each have feature vector $\pt{1}$ and probability $\origprob{1}$. Group 2 contains $n_2$ individuals that each have feature vector $\pt{2}$ and probability $\origprob{2}$.  Then, all points in Group 1 and all points in Group 2 with have the same value for $Q_i$ which is
\begin{align*}
Q_1 &= \frac{p_1(1-p_1)}{n_1},\\
Q_2 &= \frac{p_2(1-p_2)}{n_2},
\end{align*}
respectively.  From this example, we can see both the $\Theta(1/n)$ scaling of $Q_i$ and the uneven distribution of regret.  If one of the group contains much less individuals than the other, these individuals will experience higher regret.

\section{Experiments and Demonstrations}
\label{Sec::Experiments}

\subsection{Validation Studies}\label{sec:validation}

\paragraph{Setup} 

To demonstrate the effect of observational multiplicity, we use three semi-synthetic datasets.\footnote{The code for the experiments in this section is available at \url{https://github.com/ErinGeorge/Observational-Multiplicity/}.}  These are derived from the Diagnostic Wisconsin Breast Cancer dataset from \citet{wolberg1995breast}, the bank telemarketing dataset from \citet{bank-dataset}, and the sleep apnea diagnostic dataset from \citet{apnea-dataset}. We denote these datasets as \texttt{bc}, \texttt{bank}, and \texttt{apnea}, respectively.  The features are kept the same as the original datasets.  However, the labels are instead drawn according to a ground-truth logistic regression model with a fixed set of parameters.  This allows predicted regret as computed in \cref{alg:regret} to be compared with the true regret measure that would be obtained if the labels were resampled according to their true probabilities.  To ensure that a realistic relationship between features and labels is retained, the parameters for the ground-truth model is selected by training a \emph{regularized} logistic regression model to fit the original labels.  The regularization is added in this step to produce more realistic models and to avoid introducing a bias by selecting a set of parameters that are easier for unregularised logistic regression models to learn.  We consider two sets of models to test with: an unregularized logistic regression model and a random forest model.

We use \cref{alg:regret} to estimate regret using $300$ resamples of the labels.  In addition, for each experiment $100$ different trials were conducted starting from different instances of the semisynthetic dataset.  These instances differ only in which labels are originally observed.

\subsubsection{On Estimation}

\begin{figure}\centering
\includegraphics[scale=0.28]{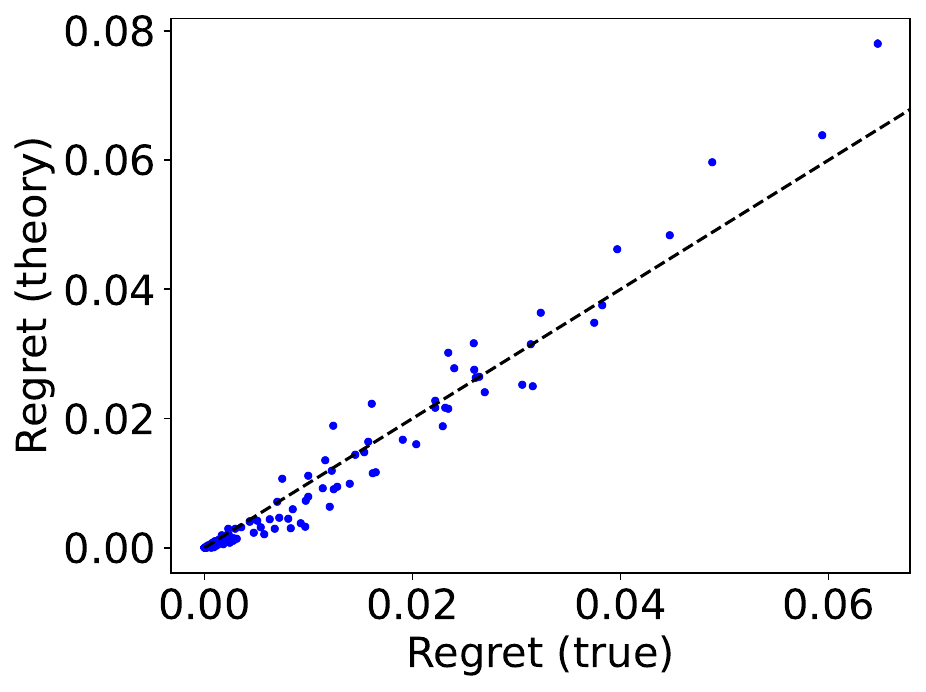}
\includegraphics[scale=0.28]{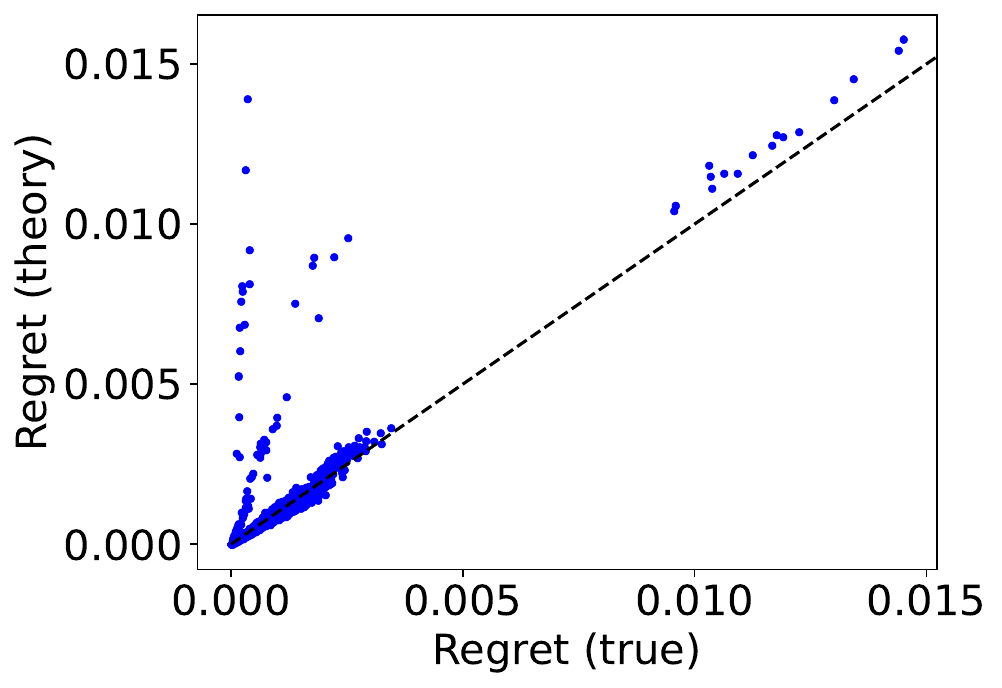}
\includegraphics[scale=0.28]{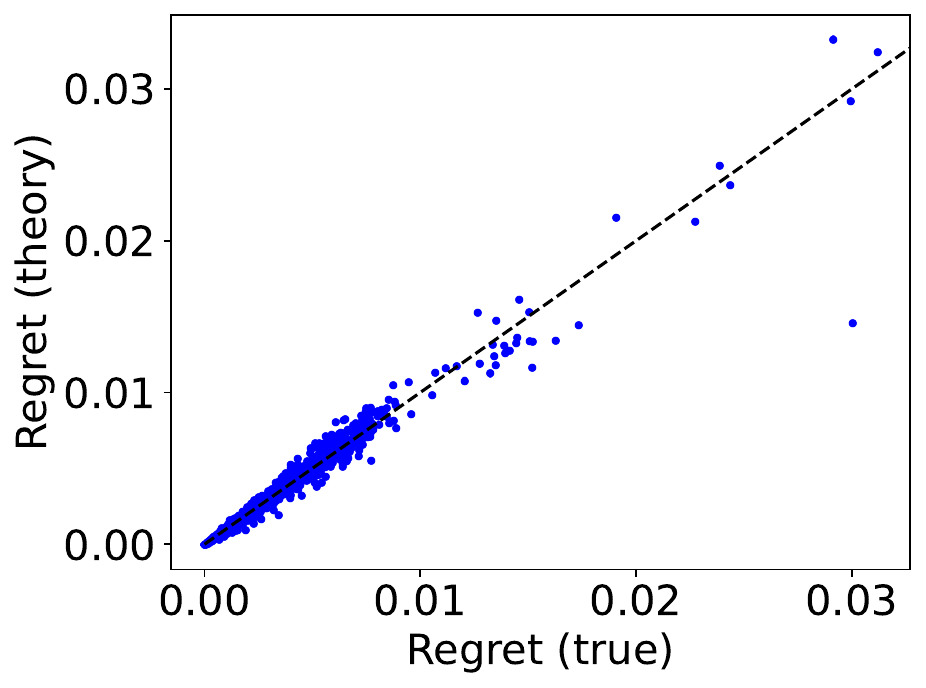}
\caption{Theorized versus actual regret in semi-synthetic datasets (from left to right: \texttt{bc}, \texttt{bank}, \texttt{apnea}) for logistic regression.  Each blue dot represents one point in the dataset and the black dashed line is the line $y=x$, representing perfect correspondence.}\label{fig:pred_vs_actual_theory}
\end{figure}

\Cref{fig:pred_vs_actual_theory} shows how the actual regret for logistic regression compares to the quantity predicted by \cref{thm:var_bd}.  We see that the predicted regret according to this theorem is generally very close to the observed regret in practice.  The vast majority of the points lie near the line $y=x$, with some notable exceptions in the \texttt{bank} dataset and one stray point in the \texttt{apnea} dataset.

\begin{figure}\centering
\includegraphics[scale=0.28]{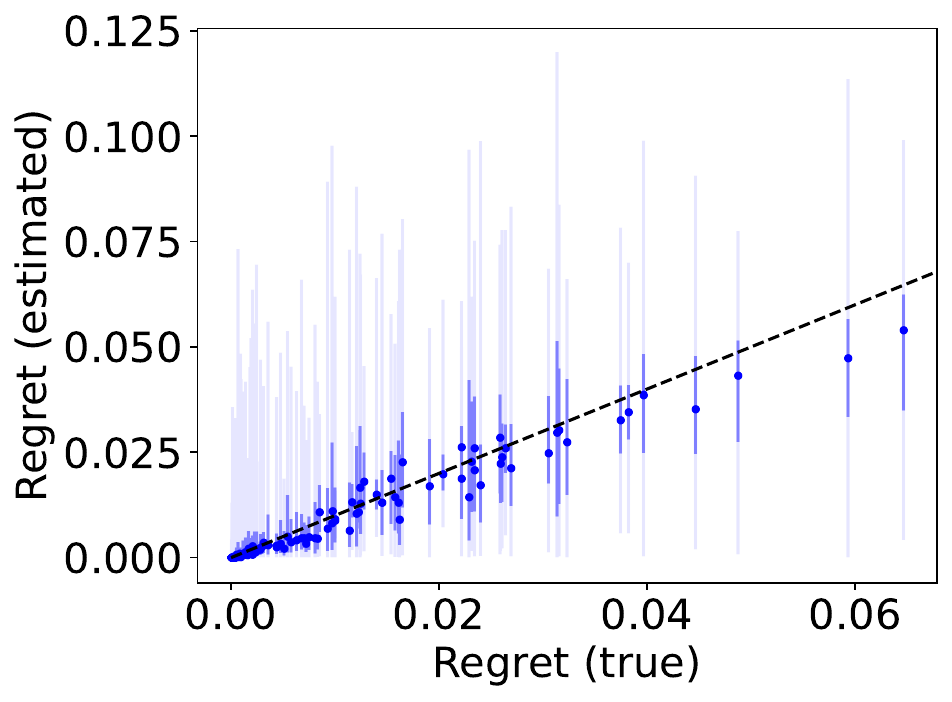}
\includegraphics[scale=0.28]{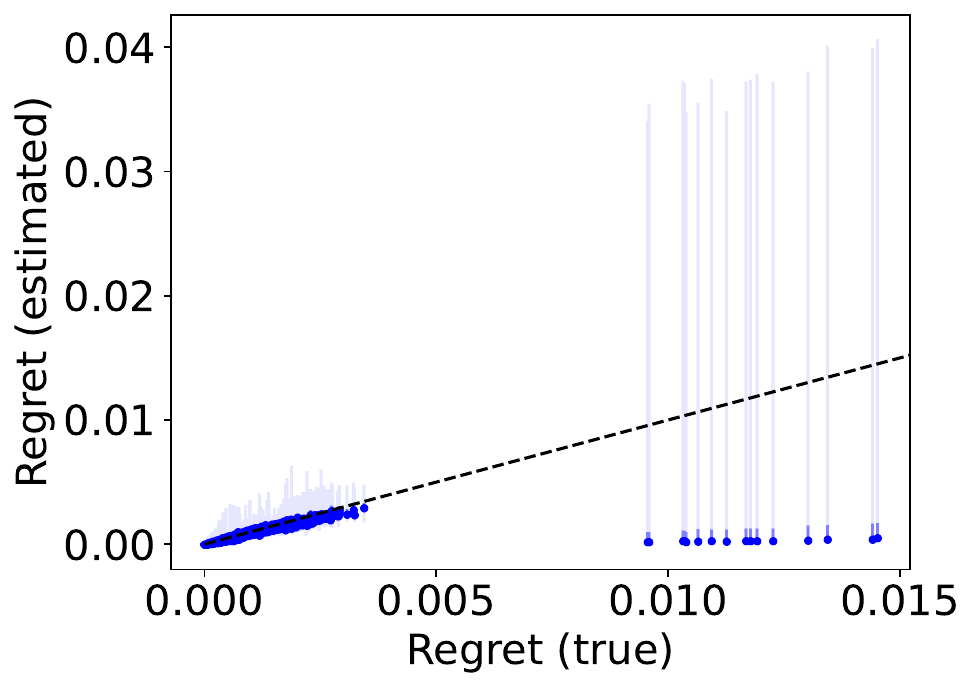}
\includegraphics[scale=0.28]{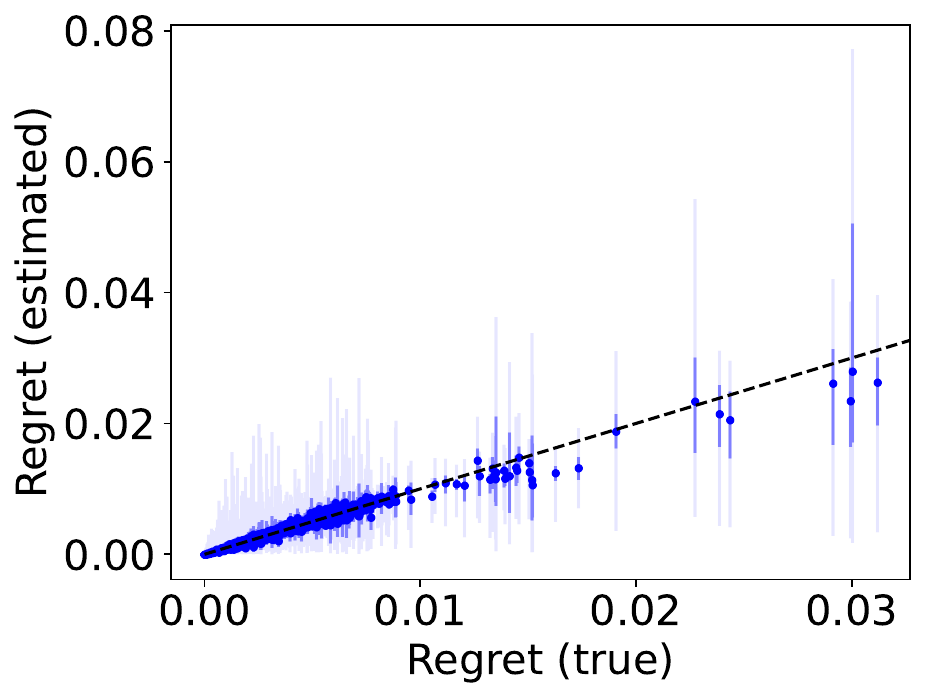}
\caption{Predicted regret versus actual regret in semi-synthetic datasets (from left to right: \texttt{bc}, \texttt{bank}, \texttt{apnea}) for logistic regression.  Each blue dot is the median of predicted regret for a point in the dataset over $100$ trials, while the dark blue and light blue lines represent the interquartile and full ranges of regrets, respectively.  The black dashed line is the line $y=x$, representing perfect prediction.}\label{fig:pred_vs_actual_exp}
\end{figure}

\begin{figure}\centering
\includegraphics[scale=0.28]{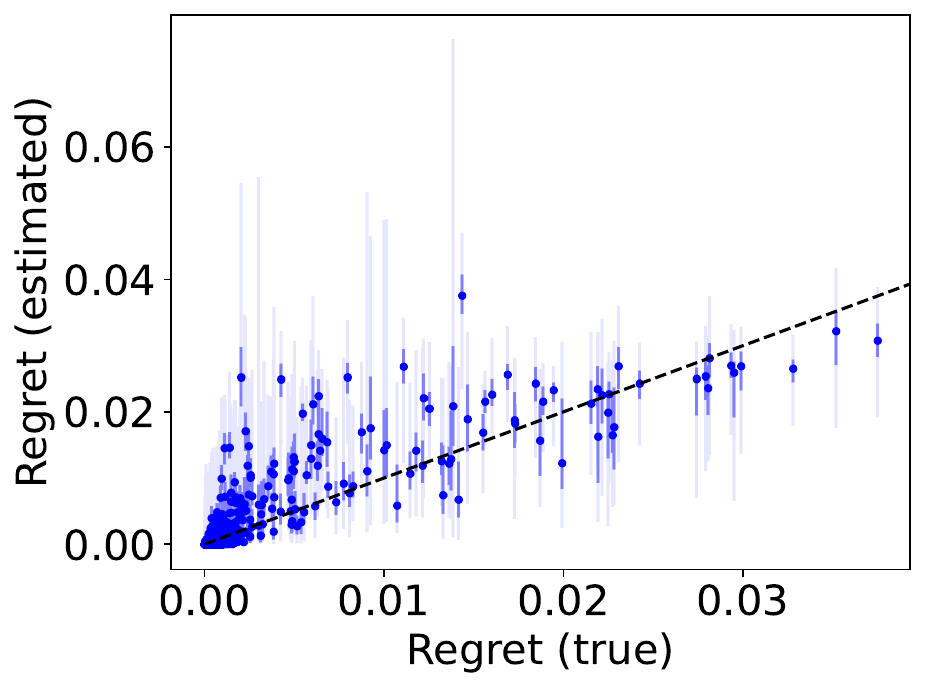}
\includegraphics[scale=0.28]{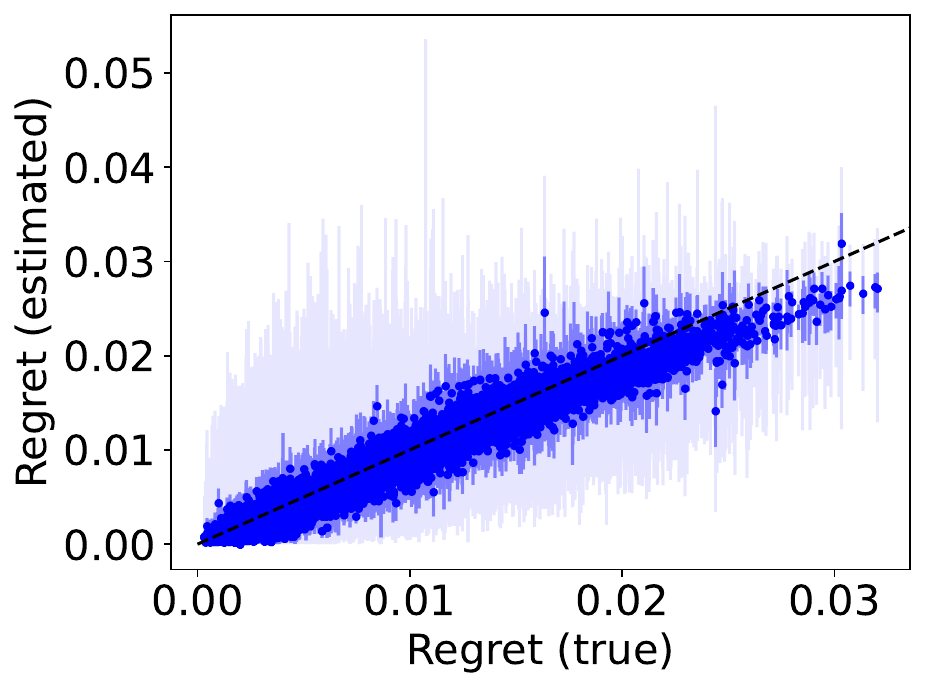}
\includegraphics[scale=0.28]{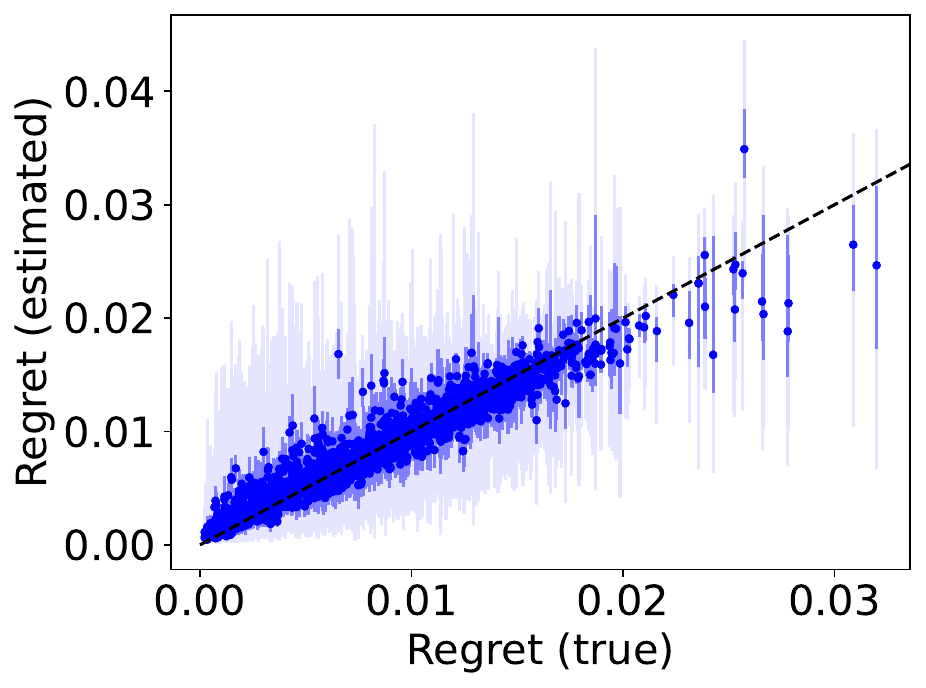}
\caption{Predicted regret versus actual regret in semi-synthetic datasets (from left to right: \texttt{bc}, \texttt{bank}, \texttt{apnea}) for random forest.  Each blue dot is the median of predicted regret for a point in the dataset over $100$ trials, while the dark blue and light blue lines represent the interquartile and full ranges of regrets, respectively.  The black dashed line is the line $y=x$, representing perfect prediction.}\label{fig:pred_vs_actual_exp_RF}
\end{figure}

\Cref{fig:pred_vs_actual_exp,fig:pred_vs_actual_exp_RF} shows the predicted regret for each point in the datasets according to \cref{alg:regret} with the true regret as the labels are resampled from their true probabilities for the logistic regression and random forest models, respectively.  For the logistic regression model, there is a large amount of variability in the estimated regret for each point, with the full range of estimated regrets being quite large in some instances.  This is particularly true in the \texttt{bc} dataset and for a few high-regret points in the \texttt{bank} dataset.  For the random forest model, the variability of predicted regret for individual points is not as high, however the predicted regrets do not cluster around the true regrets as tightly.  However, for both models there is a general linear trend and the estimated regrets concentrate around their true values over the different trials.  The main exception is the group of points in the \texttt{bank} dataset with large true regret for logistic regression.  These points typically have their regrets underestimated by \cref{alg:regret}.  We emphasize that the plots in \cref{fig:pred_vs_actual_exp} show \emph{typicality}.  It is not enough that the estimates are good on average over the $100$ different semi-synthetic labels, as in an actual use case we would only see one set of these labels.

\paragraph{\texttt{bank} dataset} The \texttt{bank} dataset exhibits behavior with the logistic regression model not present in the other two datasets.  In terms of true regret, there is a collection of outliers which exhibit much higher regret.  These points typically have their regrets underestimated by the sampling procedure, although the range of estimated regrets produced is extremely large.  Furthermore, there is an additional set of points where the true regret is much lower than what would be expected given \cref{thm:var_bd}.  These points tend to lie on two separate lines with slope much larger than one.

\begin{figure}\centering
\includegraphics[scale=0.5]{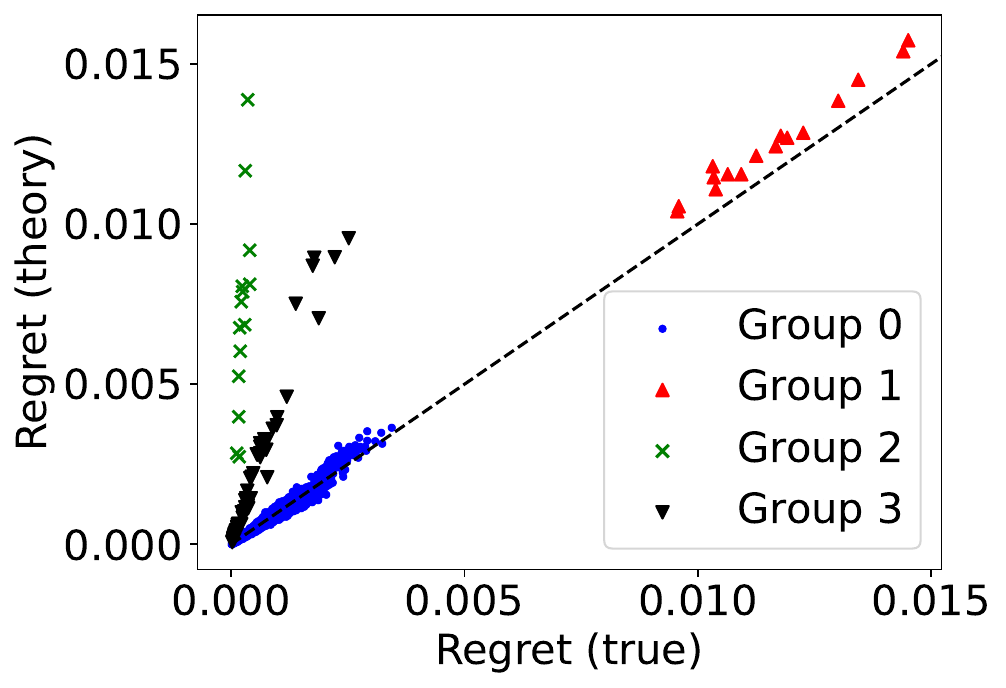}
\caption{Theorized versus actual regret in the semi-synthetic dataset \texttt{bank}.  The data is split into four groups.  Group $1$, $2$, and $3$ are determined by a specific categorical feature.  Group $0$ is the rest of the dataset.}\label{fig:pred_vs_actual_theory_group_bk}
\end{figure}

All of these groups correspond to real trends in the data itself.  The features in the \texttt{bank} dataset are all categorical variables encoded as one-hot vectors.  \Cref{fig:pred_vs_actual_theory_group_bk} shows the plot in \cref{fig:pred_vs_actual_theory} but with the data separated into four separate groups depending on the features.  Groups $1$, $2$, and $3$ are each defined by a single categorical feature and are $14$, $16$, and $80$ points, respectively.  Each of these groups consists of all the points that take a certain value in a specified feature.  Group 0 is the remaining $41{,}078$ points in the dataset.  As a result, Groups $1$, $2$, and $3$ all exist in significantly underpopulated subspaces of the data after a one-hot encoding of the categorical features, so these points are easier for a logistic regression model to identify.  It is therefore unsurprising that these points have different behavior in terms of regret compared to the full dataset.  Additionally, these underpopulated subspaces also result in the smallest eigenvalue of $\mH$ being very small.  Therefore, the condition on $\epsilon$ in \cref{thm:var_bd} is not satisified, despite the large sample size.

\subsubsection{Use Cases}

\begin{figure}\centering
\includegraphics[scale=0.27]{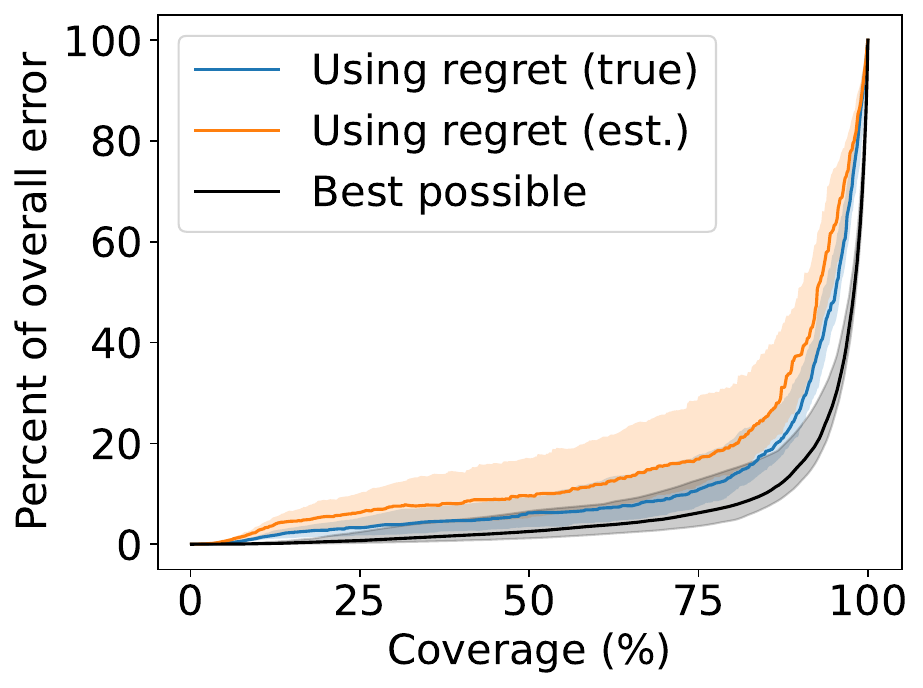}
\includegraphics[scale=0.27]{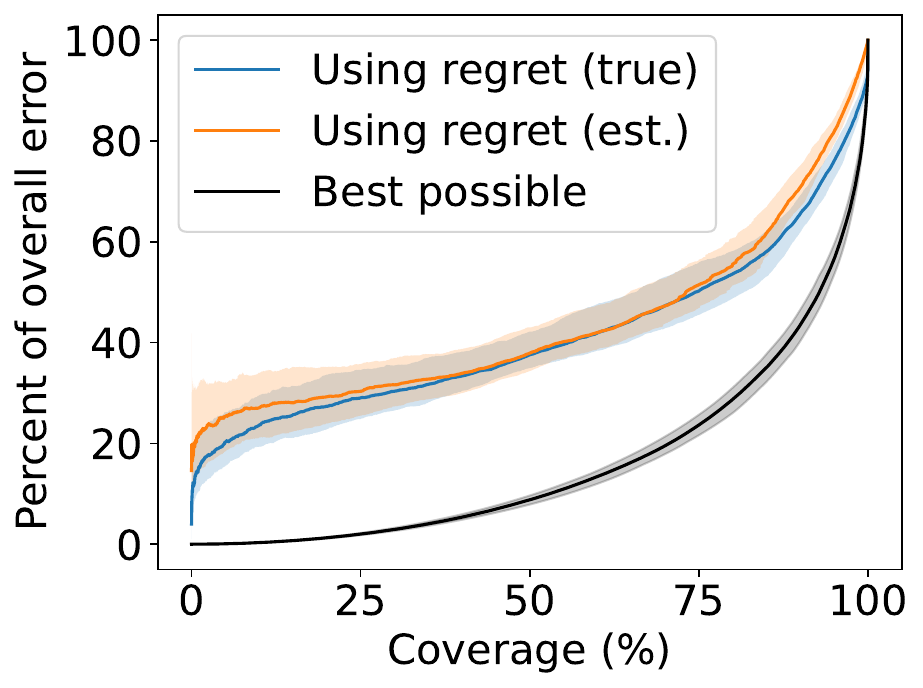}
\includegraphics[scale=0.27]{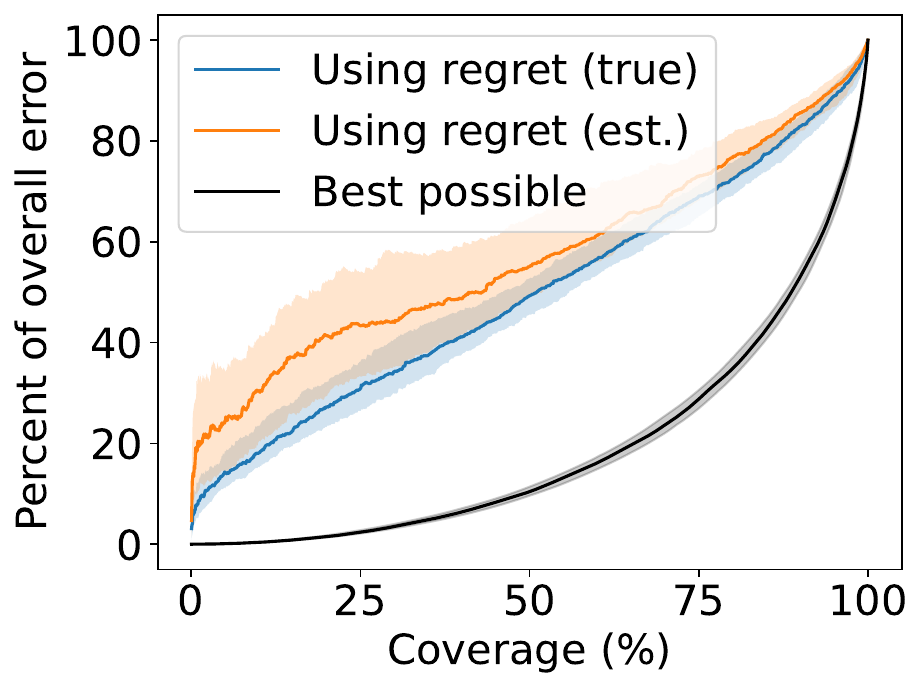}
\caption{Mean error versus coverage in selective prediction on semi-synthetic datasets (from left to right: \texttt{bc}, \texttt{bank}, \texttt{apnea}).  The median and interquartile range is shown and the selective prediction set is determined using true regret, estimated regret, and the lowest error.  The error is measured as the mean KL divergence compared to the mean KL divergence for the full dataset.}\label{fig:selective_prediction}
\end{figure}

\begin{figure}
\begin{center}
\includegraphics[scale=0.30]{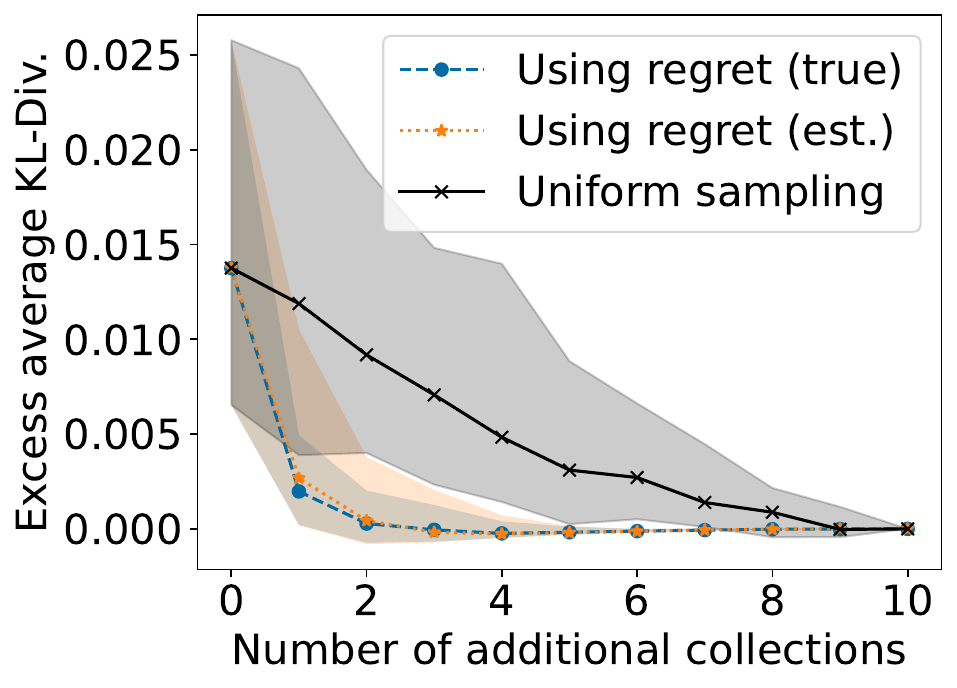}
\includegraphics[scale=0.30]{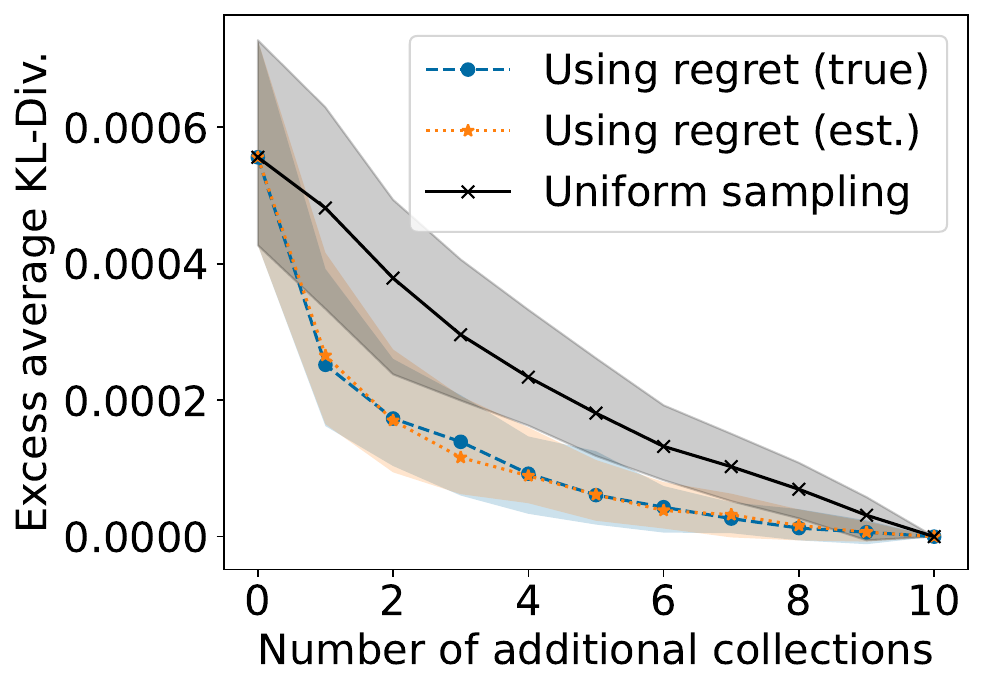}
\includegraphics[scale=0.30]{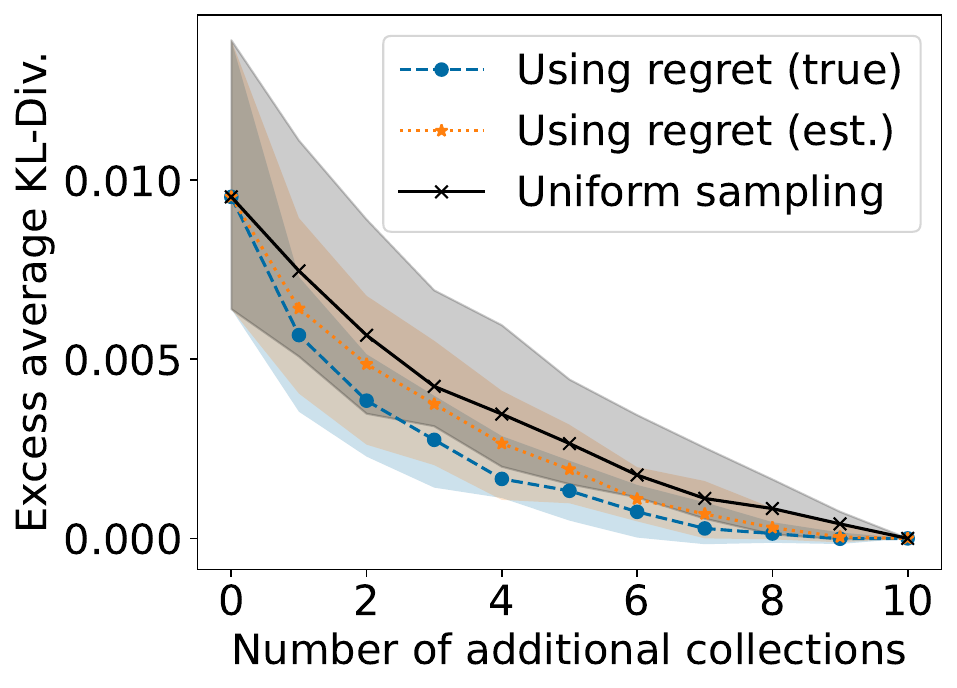}
\end{center}
\caption{Excess error in active learning learning on semi-syntehtic datasets (from left to right: \texttt{bc}, \texttt{bank}, \texttt{apnea}).  The additional collections are shown when selected according to true regret, estimated regret, and uniformly at randomly.  The median and interquartile range of excess error are shown, where excess error is the additional mean KL divergence compared to the final model on the full dataset.}\label{fig:active_learning}
\end{figure}

We now present two example use cases to show how regret can be used to improve the safety of machine learning through selective prediction and active learning, shown in \cref{fig:selective_prediction,fig:active_learning}. Regret is a measure of arbitrariness of a prediction: data points with high regret are points with predictions that are uncertain.  For these experiments, we consider only th e unregularized logistic regresion model.

\paragraph{Abstaining from arbitrary predictions} We take the learned logistic regression model and evaluate the accuracy of the model on subsets of the training dataset.  The accuracy is measured as the Kullback--Leibler divergence $D_{KL}(\natureprob{}\Vert\origprob{})$ from the predicted probabilities to the true probabilities, with the mean taken over each of the points.  All points with (true or estimated) regret below a cutoff are considered, with the cutoff varying so as to control between accuracy and coverage of the model.  This experiment is shown in \cref{fig:selective_prediction}.  We see that the accuracy of the model on the low-regret subsets of the model is typically better than the accuracy of the model of the full dataset.  Furthermore, using the estimated regret instead of the true regret results in mildly worse error and much more variation in the case of the \texttt{bc} and \texttt{apnea} datasets, but only a negligible effect in the \texttt{bank} dataset.  In the case of the \texttt{bank} dataset, using regret for selective prediction gives performance very close to the best possible, whereas for the other two datasets using regret is not as effective.

\paragraph{Guiding data collection} We start with half of the training dataset and consider a situation where we have access to the features of the other half, but not the labels.  We can observe labels for the additional points and add them to our training set.  We consider the successive performance of the model if we choose to gather information about the points with the highest regret as opposed to sampling additional points uniformed.  The results of these experiments are shown in \cref{fig:active_learning}.  Here, we see that in both the \texttt{bc} and \texttt{bank} datasets, there is significant in accuracy when using regret to determine the additional points when compared to sampling randomly.  This is most pronounced in the \texttt{bc} dataset, which gets the benefit of the full dataset after only $3$ additional samples.  Furthermore, in neither dataset is there a noticeable effect from using estimated regret instead of true regret.  This experiment is less sucessful in the \texttt{apnea} dataset, where neither true nor estimated regret are able to perform consistently better than uniform sampling, although there still appears to be a small benefit to using regret.

\subsection{Fannie Mae Single Family Loan Dataset}

In addition to our semi-synthetic experiments with logistic regression, we also compute regret for a fully real-world dataset using a gradient boosting decision tree classifier.  The dataset used was the Fannie Mae Single-Family Loan Performance Data~\cite{fannie_mae_dataset}, which we will refer to as \texttt{loan}.  We use the entries in the dataset occurring between the first quarter of 2020 and the second quarter of 2024, inclusive.  We predict whether or not a 30-day delinquency incident for a given loan is present in the dataset.  We use all features in the dataset available before a loan is given to make predictions, except for date and location derived features.  In total, there are $13{,}026{,}081$ points in the dataset and 22 features.  We take a random subset of $75\%$ of these points as our training dataset.

Computing the regret for this dataset reveals interesting phenomena.  Most points in the dataset have very low regret.  In particular, $98\%$ of the points have regret below $0.0001$.  We would expected the predicted probabilities for these points to deviate by less than a percentage point if a different dataset were observed.

\begin{figure}\centering
    \includegraphics[width=0.495\textwidth]{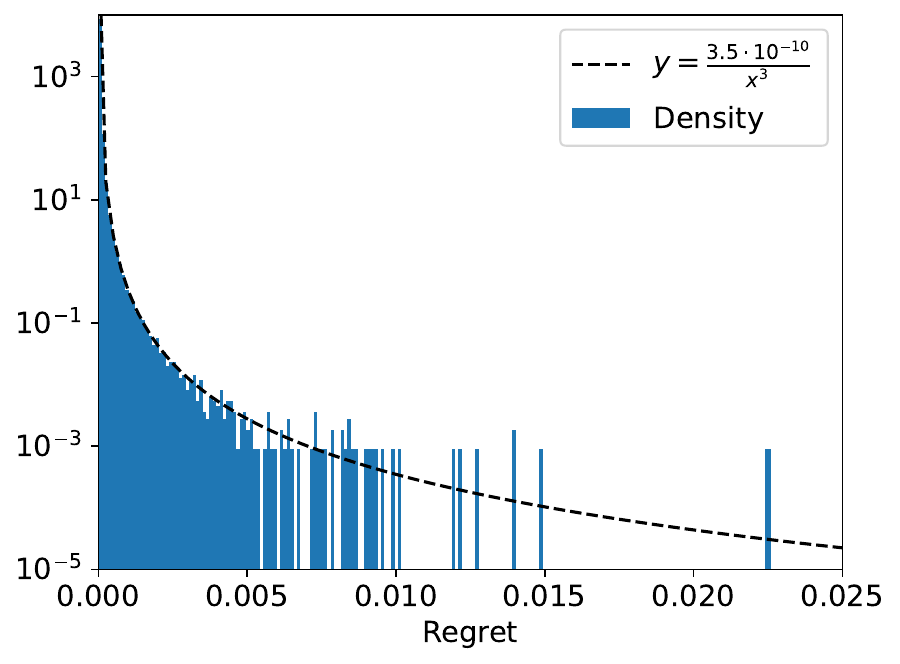}
    \includegraphics[width=0.495\textwidth]{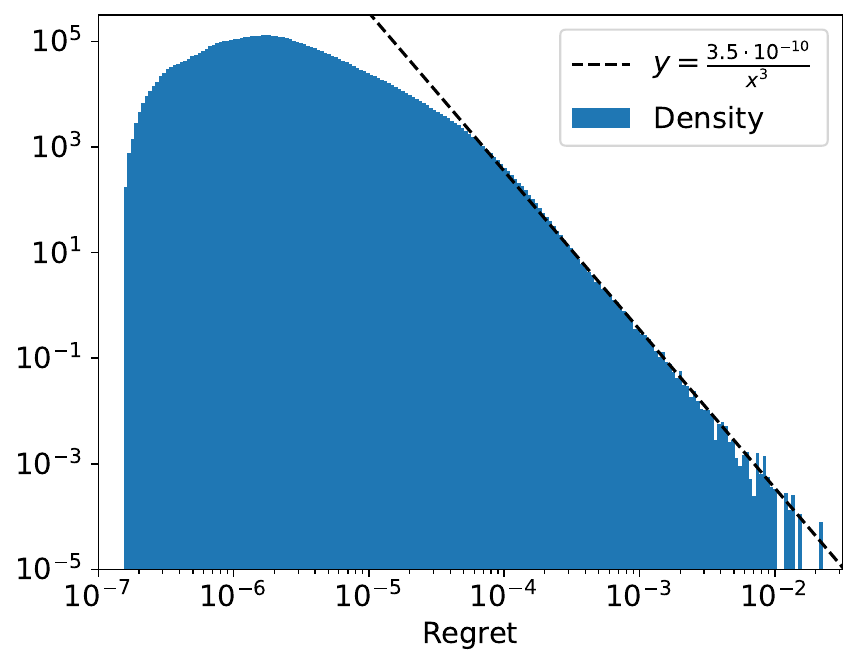}
    \caption{The distribution of the estimated regret of each point in \texttt{loan} plotted on a log scale (left) and a log-log scale (right).}\label{fig:fm-hist}
\end{figure}

However, the empirical distribution of regret in the dataset has very long tails.  This is shown in \cref{fig:fm-hist}.  The maximum regret seen in the data is $0.022$ and, in the tail of the distribution, the density appears to follow a power law.  This means that a relatively small subset of the overall population is subject to a very high regret and is likely to receive arbitrary predictions.  

\begin{figure}\centering
    \includegraphics[width=0.5\textwidth]{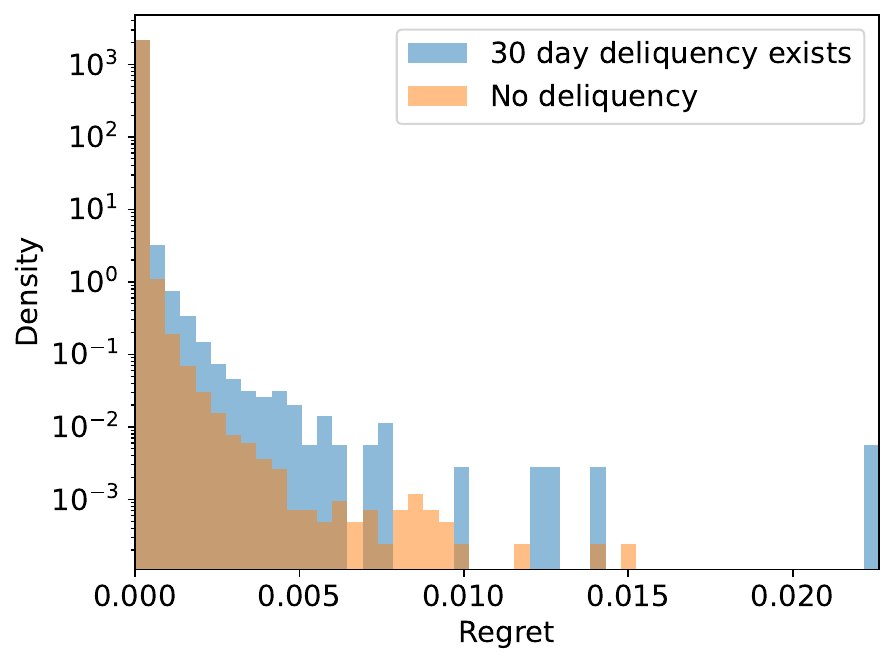}
    \caption{The distribution of the estimated regret for each point in \texttt{loan} on a log scale, separated by observed training label.}\label{fig:fm-hist-label}
\end{figure}

An immediate question to consider: what does this subset of people look like in comparison to the population at large?  \Cref{fig:fm-hist-label} shows the distribution of regret in the training dataset depending on the initially observed training label.  Individuals who had a recorded delinquency are more likely to have high regret and, conversely, those with high regret are more likely to have a recorded delinquency.

\section{Analysis of Logistic Regression}\label{sec:proof}

We now turn to prove the results stated in \cref{sec:theorems}.  First, we introduce the mathematical setting and notation we will be using for the proofs.

\subsection{Preliminaries}

We consider a logistic regression task.  When we run ERM on a resampled dataset $\resampdata$, it returns the unique best-fit model (if it exists) with the parameters:
\begin{align*}
    \resampparam \defeq \argmin_{\paramvec \in \R^d} \resamploss{\paramvec} = \argmin_{\paramvec \in \R^d} \sum_{i=1}^n \log \big(1 + \exp(-\resamplabel{i} \pt{i}^\transpose \paramvec)\big)\numberthis\label{eq:argmin_theta}
\end{align*}
where $\resamploss{\paramvec}$ denotes the logistic loss at parameter $\paramvec$ on the dataset $\resampdata$.  If \cref{eq:argmin_theta} does not admit a unique solution, we will allow $\resampparam$ to take any value in $\R^d$.  We will see later that the probability of this occurring is small in the setting we are working with, so this is not an issue.

Given the model, we can compute the predicted probability for an instance: \[\resampprob{i} = \mathrm{Pr}(\resamplabel{i} = 1 \;|\; \pt{i}) = \frac{1}{1+\exp(-\pt{i}^\transpose \resampparam)}.\]
The expression for $\origprob{i}$ is $\sigma(\pt{i}^\transpose\resampparam)$, where $\sigma$ is the sigmoid function.  The sigmoid function is defined by
\[\sigma(s) \defeq \frac{1}{1+\exp(-s)}.\]
Useful properties of $\sigma$ include
\begin{align*}
    \sigma(-s) &= 1-\sigma(s) \\
    \sigma'(s) &= \sigma(s)(1-\sigma(s)).
\end{align*}
Recall that we constructed our dataset $\resampdata$ by sampling a set of $n$ \emph{plausible labels}, one for each point $\pt{i}$, as follows:
\begin{align}
\resamplabel{i} = \begin{cases}1 &\text{ with probability $\origprob{i}$} \\ -1 &\text{ with probability $1-\origprob{i}$,}\end{cases}
\end{align}
where $\origprob{i} = \sigma(\pt{i}^\transpose\origparam)$ for some parameter vector $\origparam\in\R^d$.
We wish to study how the predicted probability for each data point changes with this model.   To do so, we consider the derivative of the loss at $\origparam$.  This is as follows:
\begin{align*}
    \nabla_{\paramvec}\resamploss{\origparam} &= \sum_{i=1}^n -\resamplabel{i}\pt{i}\sigma(-\resamplabel{i}\pt{i}^\transpose \origparam)\\
    &= \sum_{i=1}^n a_i \pt{i},
\end{align*}
where $a_i$ is defined as
\begin{align*}
a_i = \begin{cases}-(1-\origprob{i}) &\text{if $\resamplabel{i} = 1$ (with probability $\origprob{i}$)}\\\origprob{i} &\text{if $\resamplabel{i} = -1$ (with probability $1-\origprob{i}$).} \end{cases}
\end{align*}
The random variables $a_i$ are all independent and satisfy
\begin{align}
    \expec(a_i) &= 0 \label{eq:ai_ev}\\
    \expec(a_i^2) &= \origprob{i}(1-\origprob{i}). \label{eq:ai2_ev}
\end{align}

\subsubsection{Notation}

We will now introduce the following notation to analyze the problem.  For a vector $\pt{}$, we denote the $\ell_2$-norm by $\|\pt{}\|_2$.  Let $\grad \defeq \nabla_{\paramvec}\resamploss{\origparam}$ be the derivative of the loss at $\origparam$.  Let
\[\Hess \defeq \nabla^2_{\paramvec} \resamploss{\origparam} = \sum_{i=1}^n \origprob{i}(1-\origprob{i}) \pt{i} \pt{i}^\transpose\]
be the Hessian of the loss at $\origparam$.  Note that this expression does not depend on the resampled labels.  Let $\eigmin$ and $\eigmax$ be the minimum and maximum eigenvalues of $\Hess$, respectively.  Let $\xmax \defeq \max_i \|\pt{i}\|_2$ and $\xmin \defeq \min_i \|\pt{i}\|_2$.  We define
\begin{align*}
\innerHinv{\bm{u}}{\bm{v}} &\defeq \bm{u}^\transpose \invHess \bm{v},\\
\normHinv{\bm{u}} &\defeq \sqrt{\innerHinv{\bm{u}}{\bm{u}}}
\end{align*}
for $\bm{u}, \bm{v} \in \R^d$.  Note that because we assume our data is full rank, $\Hess$ is positive definite.  This means $\invHess$ is also positive definite, which ensures $\innerHinv{\cdot}{\cdot}$ is an inner product.

It will be useful to consider the event $\eventsmall{c}$ that occurs whenever $\normHinv{\grad}<c$.  We will also consider $\eventsmallcomp{c}$, which is the event that this does not happen.  We use $\ind(\cdot)$ to be the indicator function of an event, which is $1$ when the event occurs and $0$ otherwise.

\subsubsection{Proof sketch}

A quick sketch of the proof of the theorem is as follows.  From a result of \citet{bach2010self}, we see that $\resampparam-\origparam$ is approximately $-\invHess\grad$ provided $\normHinv{\grad}$ is not too large.  Under this condition, $\pt{i}^\transpose(\resampparam-\origparam)$ is approximately $-\pt{i}^\transpose\invHess\grad$.  This last expression admits simple formulas for its first two moments.  We complete the proof by bounding the probability that $\normHinv{\grad}$ is large and taking a Taylor approximation to bound the moments of $\sigma\big(\pt{i}^\transpose(\resampparam-\origparam)\big)$.

\subsection{Initial lemmas}

To prove \cref{thm:var_bd}, we will make use of the following result, restated in the context we are working in:
\begin{theorem}[\citet{bach2010self}, Proposition 2]\label{thm:NM_bd}
    If $\normHinv{\grad} \leq \sqrt{\eigmin}/(2\xmax)$, then $\resamploss{\paramvec}$ has a finite global minimum $\resampparam$ and the following hold:
    \begin{align*}
        \normHinv{\Hess(\origparam - \resampparam)} &\leq 4\normHinv{\grad},\\
        \normHinv{\Hess(\origparam - \invHess\grad -\resampparam)} &\leq \frac{4\xmax}{\sqrt{\eigmin}}\normHinv{\grad}^2.
    \end{align*}
\end{theorem}
This result is a quantitative bound on how closely Newton's method in optimization approximates the global minimum of logistic regression in a single step.  Consequently, to analyze the distribution of $\pt{i}^\transpose(\resampparam - \origparam)$, it suffices to consider $\innerHinv{\pt{i}}{\grad}$ and bound $\normHinv{\grad}$.  The following lemma bounds the latter with high probability:

\begin{lemma}\label{lem:grad_bd}
    For $c < \sqrt{\eigmin}/(2\xmax)$, we have
\begin{align*}
   \prob\left(\eventsmallcomp{c}\right) \leq 2d\exp\left(-\frac{c^2}{3d}\right).
\end{align*}
\end{lemma}

\begin{proof}
We use Bernstein's inequality.  Let $\bm{u}_1, \ldots, \bm{u}_d$ be an orthonormal basis for $\R^d$ with respect to $\innerHinv{\cdot}{\cdot}$.  Then, for any constant $c$, $\normHinv{\grad}^2 \geq c^2$ only if
\[|\innerHinv{\grad}{\bm{u}_i}| \geq \frac{c}{\sqrt{d}}\]
for some $i$.  We can write
\begin{align*}
   \innerHinv{\grad}{\bm{u}_i} 
   &= \sum_{j=1}^n a_j \innerHinv{\pt{j}}{\bm{u}_i},
\end{align*}
which is the sum of $n$ zero-mean random variables that take values in
\[[-\xmax\|\invHess\bm{u}_i\|_2, \xmax\|\invHess\bm{u}_i\|_2].\]
Note that
\[\|\invHess\bm{u}_i\|_2 \leq \frac{1}{\sqrt{\eigmin}}\innerHinv{\bm{u}_i}{\bm{u}_i} = \frac{1}{\sqrt{\eigmin}}.\] We also see that
\[\sum_{j=1}^n \expec\big((a_j\innerHinv{\pt{j}}{\bm{u}_i})^2\big) = \sum_{j=1}^n \bm{u}_i^\transpose (\invHess)^\transpose\origprob{j}(1-\origprob{j})\pt{j}\pt{j}^\transpose\invHess\bm{u}_i = \innerHinv{\bm{u}_i}{\bm{u}_i}  = 1\]
By Bernstein's inequality, for a fixed $i$,
\begin{align*}
   \prob\left(|\innerHinv{\grad}{\bm{u}_i}| \geq \frac{c}{\sqrt{d}}\right) \leq 2 \exp\left(-\frac{c^2/(2d)}{1 + c\xmax/(3\sqrt{d\eigmin})}\right).
\end{align*}
By a union bound and the assumption $c < \sqrt{\eigmin}/(2\xmax)$, we obtain
\begin{equation*}
   \prob\left(\eventsmallcomp{c}\right) \leq 2d \exp\left(-\frac{c^2/(2d)}{1 + 1/(6\sqrt{d})}\right) \leq 2d\exp\left(-\frac{c^2}{3d}\right).\qedhere
\end{equation*}
\end{proof}

\subsection{Moments of logits}

We now proceed to demonstrate the following lemma, which bounds the first two moments of the logit probability (the input to $\sigma$) when the event $\eventsmall{c}$ holds.

\begin{lemma}\label{lem:logit-bds}
Define $Z_i$ to be the random variable $\pt{i}^\transpose(\resampparam-\origparam)$.  For any real number $c < \sqrt{\eigmin}/(2\xmax)$, we have the following bounds 
\begin{align}
\left|\sqrt{\expec\big(Z_i^2\cdot\ind(\eventsmall{c})\big)} -\normHinv{\pt{i}}\right|&\leq \normHinv{\pt{i}}\left(\sqrt{2dn} \exp\left(\|\origparam\|_2\xmax-\frac{c^2}{6d}\right)+\frac{4c^2\xmax}{\sqrt{\eigmin}}\right)\label{eq:logitsq_ub}, \\
|\expec\big(Z_i\cdot\ind(\eventsmall{c})\big)| &\leq \normHinv{\pt{i}} \left(2d\sqrt{n}\exp\left(\|\origparam\|_2\xmax-\frac{c^2}{3d}\right) + \frac{4c^2\xmax}{\sqrt{\eigmin}}\right).\label{eq:logitmean_ub}
\end{align}
\end{lemma}
\begin{proof}
First, we demonstrate \cref{eq:logitsq_ub}.  We start with the following identity:
\[\expec(\grad{\grad}^\transpose) = \sum_{i=1}^n\sum_{j=1}^n \expec(a_ia_j)\pt{i}\pt{j}^\transpose = \sum_{i=1}^n \expec(a_i^2)\pt{i}\pt{i}^\transpose = \sum_{i=1}^n \origprob{i}(1-\origprob{i})\pt{i}\pt{i}^\transpose = \Hess.\]
From this, we derive a second identity:
\begin{align*}
    \expec(\innerHinv{\pt{i}}{\grad}^2) =\expec(\pt{i}^\transpose \invHess \grad{\grad}^\transpose \invHess\pt{i}) = \normHinv{ \pt{i}}^2.\numberthis\label{eq:SQ-expec-xi-dot-invgradhess-bd}
\end{align*}
This allows us to write the left hand side of \cref{eq:logitsq_ub} as
\[\sqrt{\expec\big(Z_i^2\cdot\ind(\eventsmall{c})\big)} -\normHinv{\pt{i}} = \sqrt{\expec\big(Z_i^2\cdot\ind(\eventsmall{c})\big)} -\expec(\innerHinv{\pt{i}}{\grad}^2).\]
We then use triangle inequality:
\begin{align*}
\begin{split}
    \left|\sqrt{\expec\big(Z_i^2 \cdot\ind(\eventsmall{c})\big)}-\normHinv{\pt{i}}\right| \leq&    \left|\sqrt{\expec\big(Z_i^2 \cdot\ind(\eventsmall{c})\big)}-\sqrt{\expec\big(\innerHinv{\pt{i}}{\grad}^2 \cdot\ind(\eventsmall{c})\big)}\right|\\
    &\qquad +\left|\sqrt{\expec\big(\innerHinv{\pt{i}}{\grad}^2 \cdot\ind(\eventsmall{c})\big)} - \sqrt{\expec(\innerHinv{\pt{i}}{\grad}^2 )}\right|
\end{split}\\
\begin{split}
    \leq& \sqrt{\expec\bigg(\big(\pt{i}^\transpose(\resampparam - \origparam + \invHess \grad)\big)^2 \cdot\ind(\eventsmall{c})\bigg)}\\
    &\qquad + \sqrt{\expec\big(\innerHinv{\pt{i}}{\grad}^2 \cdot\ind(\eventsmallcomp{c})\big)}.
\end{split}
\numberthis\label{eq:logitsquare-triangle-ineq}
\end{align*}
Using \cref{thm:NM_bd}, it is immediate to bound the first term on the right hand side of \cref{eq:logitsquare-triangle-ineq}:
\begin{align*}
        \left|\pt{i}^\transpose(\resampparam - \origparam + \invHess  \grad)\right|\cdot\ind(\eventsmall{c})  &= |\innerHinv{\pt{i}}{\Hess(\origparam -\invHess\grad - \resampparam )}|\cdot\ind(\eventsmall{c}) \\
    &\leq \frac{4c^2\xmax}{\sqrt{\eigmin}}\normHinv{\pt{i}}\numberthis\label{eq:NM_err}.
\end{align*}

We now consider the term $\sqrt{\expec\big(\innerHinv{\pt{i}}{\grad}^2 \cdot\ind(\eventsmall{c})\big)}$ from \cref{eq:logitsquare-triangle-ineq}.  To bound this, we need an upper bound for $|\innerHinv{\pt{i}}{\grad}|$ that is valid with full probability.  This can be obtained as follows:
\begin{align*}
    \left|\innerHinv{\pt{i}}{\grad}\right| 
    &\leq \sum_{j=1}^{n} \left| a_j \pt{i}^\transpose \invHess \pt{j} \right| \\
    &\leq \frac{1}{\min_j\min\{\origprob{j},1-\origprob{j}\}}\sum_{j=1}^n \origprob{j}(1-\origprob{j}) |\pt{i}^\transpose \invHess \pt{j}| \\
    &\leq \left(1+\exp(\|\origparam\|_2\xmax) \right)\sqrt{\sum_{j=1}^n \origprob{j}(1-\origprob{j})} \sqrt{\sum_{j=1}^n \origprob{j}(1-\origprob{j})(\pt{i}^\transpose \invHess \pt{j})^2} \\
    &\leq \sqrt{n}\exp(\|\origparam\|_2\xmax)\sqrt{\sum_{j=1}^n \pt{i}^\transpose \invHess\origprob{j}(1-\origprob{j}) \pt{j}\pt{j}^\transpose \invHess \pt{i}} \\
    &= \normHinv{ \pt{i}} \cdot \sqrt{n}\exp(\|\origparam\|_2\xmax).\numberthis\label{eq:fullprob-grad-bd}
\end{align*}
This yields the bound
\begin{align*}
    \sqrt{\expec\big(\innerHinv{\pt{i}}{\grad}^2 \cdot\ind(\eventsmall{c})\big)} &\leq \normHinv{\pt{i}}\cdot \sqrt{n\prob\left(\eventsmallcomp{c}\right)} \exp(\|\origparam\|_2\xmax).
\end{align*}
After we substituting in the bound for $\prob(\eventsmallcomp{c})$ from \cref{lem:grad_bd}, we have
\begin{align*}
    \sqrt{\expec\big(\innerHinv{\pt{i}}{\grad}^2 \cdot\ind(\eventsmall{c})\big)} &\leq \normHinv{\pt{i}}\cdot \sqrt{2dn} \exp\left(\|\origparam\|_2\xmax-\frac{c^2}{6d}\right).\numberthis\label{eq:sq-badevent-bd}
\end{align*}
\Cref{eq:logitsq_ub} follows from \cref{eq:logitsquare-triangle-ineq,eq:NM_err,eq:sq-badevent-bd}.

We now proceed to demonstrate \cref{eq:logitmean_ub}.  We again use the triangle inequality and \cref{eq:NM_err} to bound:
\begin{align*}
\begin{split}
|\expec\big(\pt{i}^\transpose(\resampparam-\origparam) \cdot\ind(\eventsmall{c})\big)| \leq& \left|\expec\big(\innerHinv{\pt{i}}{\grad} \cdot\ind(\eventsmall{c})\big)\right| \\
    &\qquad + \left|\expec\big(\pt{i}^\transpose(\resampparam - \origparam + \invHess \grad)\cdot\ind(\eventsmall{c})\big)\right|
    \end{split}\\
    \leq& \left|\expec\big(\innerHinv{\pt{i}}{\grad}\cdot\ind(\eventsmall{c})\big)\right| + \frac{4c^2\xmax}{\sqrt{\eigmin}}\normHinv{\pt{i}}\numberthis\label{eq:EV-changelogitcond-ub}
\end{align*}
We have that $\expec(\invHess\grad)=0$.  Therefore, the first term on the right hand side of \cref{eq:EV-changelogitcond-ub} can be bounded by
\begin{align*}
    \left|\expec\big(\innerHinv{\pt{i}}{\grad} \cdot\ind(\eventsmall{c})\big)\right| &= \left|\expec(\innerHinv{\pt{i}}{\grad}) - \expec\big(\innerHinv{\pt{i}}{\grad} \cdot\ind(\eventsmallcomp{c})\big)\right| \\
    &= \left|\prob(\eventsmallcomp{c})\expec\big(\innerHinv{\pt{i}}{\grad}\;\mid\;\eventsmallcomp{c})\big)\right| \\
    &\leq \sqrt{n}\prob(\eventsmallcomp{c})\exp(\|\origparam\|_2\xmax)\normHinv{ \pt{i}}\numberthis\label{eq:EV-xi-dot-invhess-cond-ub},
\end{align*}
using \cref{eq:fullprob-grad-bd}.  \Cref{eq:logitmean_ub} follows from \cref{lem:grad_bd,eq:EV-changelogitcond-ub,eq:EV-xi-dot-invhess-cond-ub}.
\end{proof}

\subsection{\texorpdfstring{Proof of \cref{thm:var_bd}}{Proof of Theorem~\ref{thm:var_bd}}}
We are now ready to restate \cref{thm:var_bd}.
\thmlogregprob*

\begin{proof}
    We use the formula
\begin{align*}
    \Var(\resampprob{i}) = \Var(\resampprob{i}-\origprob{i}) = \expec\big((\resampprob{i}-\origprob{i})^2\big) - \big(\expec(\resampprob{i}-\origprob{i})\big)^2\numberthis\label{eq:var_decomp}
\end{align*}
and bound the two terms on the right hand side.  Let $c$ be a real number to be defined in full later.  For now, we will impose $c < \sqrt{\eigmin}/(40\xmax)$ to ensure the conditions of prior lemmas are met and to assist in optimizing the constant for $\epsilon$.  We will use a Taylor expansion of $\sigma$ about $\pt{i}^\transpose\origparam$ and apply \cref{lem:logit-bds}.  By \cref{thm:NM_bd}, whenever $\eventsmall{c}$ holds we have
\begin{align*}
    \left|\pt{i}^\transpose(\origparam-\resampparam)\right| &= |\innerHinv{\pt{i}}{\Hess(\origparam-\resampparam)}| \\
    &\leq \normHinv{\pt{i}}\left(\normHinv{\Hess(\origparam-\invHess\grad-\resampparam)}+\normHinv{\grad}\right) \\
    &\leq \normHinv{\pt{i}}\left(\frac{4c^2\xmax}{\sqrt{\eigmin}}+ c\right).
\end{align*}
Therefore, whenever $\eventsmall{c}$ holds, we can find a $t_*$ with $|t_*|\leq 1.1c\normHinv{\pt{i}}$ such that
\begin{align*}
    \resampprob{i} = \origprob{i} + \sigma'(\pt{i}^\transpose\origparam)\pt{i}^\transpose(\resampparam-\origparam) + \frac{1}{2}\sigma''(\pt{i}^\transpose\origparam+t_*)\big(\pt{i}^\transpose(\resampparam-\origparam)\big)^2.
\end{align*}
Recall that $\sigma'(\pt{i}^\transpose\origparam) = \origprob{i}(1-\origprob{i})$.  Consequently, if we let \[R_i \defeq \resampprob{i} - \origprob{i} - \origprob{i}(1-\origprob{i})\pt{i}^\transpose(\resampparam-\origparam),\]
then
\begin{align*}
    |R_i|\cdot\ind(\eventsmall{c}) &\leq \frac{1.1^2}{2}c^2\normHinv{\pt{i}}^2\cdot\left(\max_{|t|\leq 4c\normHinv{\pt{i}}}|\sigma''(\pt{i}^\transpose\origparam+t)|\right)
\end{align*}
almost surely.  The second derivative of $\sigma$ satisfies
\[|\sigma''(s)| = \left|\sigma(s)\big(1-\sigma(s)\big)\big(1-2\sigma(s)\big)\right| \leq |\sigma(s)\sigma(-s)|.\]
This bound on the right hand side is symmetric about $s=0$, attains its maximum at $s=0$, and strictly decreasing for $s \geq 0$.  Hence, there is $s_* \leq 1.1c\normHinv{\pt{i}}$ such that
\begin{align*}
    \max_{|t|\leq 4c\normHinv{\pt{i}}}|\sigma''(\pt{i}^\transpose\origparam+t)| &\leq \sigma(|\pt{i}^\transpose \origparam|-s_*)\sigma(-|\pt{i}^\transpose\origparam|+s_*) \\
    &\leq \sigma(|\pt{i}^\transpose\origparam|)\sigma(-|\pt{i}^\transpose|+s_*) \\
    &\leq \exp(s_*)\sigma(|\pt{i}^\transpose\origparam|)\sigma(-|\pt{i}^\transpose\origparam|).
\end{align*}
From this computation, the bound $\normHinv{\pt{i}}\leq \xmax/\sqrt{\eigmin}$, and our imposed condition on $c$, we have the almost sure bound
\begin{align*}
    |R_i|\cdot \ind(\eventsmall{c}) 
    &\leq \sqrt{Q_i}\cdot\frac{1.14c^2\xmax}{\sqrt{\eigmin}}.\numberthis\label{eq:Taylor_err}
\end{align*}
The constant in front comes from the bound $1.21\exp(1.1/40)/2 \leq 1.14$.
With the remainder from the Taylor approximation bounded, we return to \cref{eq:var_decomp}.  Of the two terms to bound in this equation, we first consider $\big(\expec(\resampprob{i}-\origprob{i})\big)^2$.  We bound
\begin{align*}
    |\expec(\resampprob{i}-\origprob{i})| &\leq |\expec\big((\resampprob{i}-\origprob{i})\cdot\ind(\eventsmall{c})\big)| + \prob(\eventsmallcomp{c}) \\
    &\leq |\expec\big(\origprob{i}(1-\origprob{i})\pt{i}^\transpose(\resampparam-\origparam)\cdot\ind(\eventsmall{c})\big)| + \expec\big(|R_i|\cdot\ind(\eventsmall{c})\big) + \prob(\eventsmall{c}).
\end{align*}
We now apply \cref{lem:grad_bd,lem:logit-bds,eq:Taylor_err} to conclude
\begin{align*}
\begin{split}
    |\expec(\resampprob{i}-\origprob{i})| \leq&
    \sqrt{Q_i}\left(2d\sqrt{n}\exp\left(\|\origparam\|_2\xmax-\frac{c^2}{3d}\right) + \frac{5.54c^2\xmax}{\sqrt{\eigmin}}\right) \\
    &\qquad +2d\exp\left(-\frac{c^2}{3d}\right).
    \end{split}
    \numberthis\label{eq:pi_1st_moment_bd}
\end{align*}
Now consider $\expec\big((\resampprob{i}-\origprob{i})^2\big)$ from \cref{eq:var_decomp}.  We bound
\begin{align*}
    \left|\sqrt{\expec\big((\resampprob{i}-\origprob{i})^2\big)}-\sqrt{Q_i}\right| \leq& \left|\sqrt{\expec\big((\resampprob{i}-\origprob{i})^2\cdot\ind(\eventsmall{c})\big)}-\sqrt{Q_i}\right| + \prob(\eventsmall{c}) \\
    \begin{split}
    \leq&\left|\sqrt{\expec\bigg(\origprob{i}^2(1-\origprob{i})^2\big(\pt{i}^\transpose(\resampparam-\origparam)\big)^2\cdot\ind(\eventsmall{c})\bigg)}-\sqrt{Q_i}\right| \\
&\qquad + \sqrt{\expec\big(R_i^2\cdot\ind(\eventsmall{c})\big)}+\prob(\eventsmall{c}) 
    \end{split}
\end{align*}
We again apply \cref{lem:grad_bd,lem:logit-bds,eq:Taylor_err} to conclude
\begin{align*}
\begin{split}
    \left|\sqrt{\expec\big((\resampprob{i}-\origprob{i})^2\big)}-\sqrt{Q_i}\right| \leq& \sqrt{Q_i}\left(\sqrt{2dn}\exp\left(\|\origparam\|_2\xmax-\frac{c^2}{6d}\right)+\frac{5.54c^2\xmax}{\sqrt{\eigmin}}\right) \\
    &\qquad + 2d\exp\left(-\frac{c^2}{3d}\right).
\end{split}\numberthis\label{eq:pi_2nd_moment_bd}
\end{align*}
\Cref{eq:pi_1st_moment_bd,eq:pi_2nd_moment_bd} simplify as follows:
\begin{align*}
    |\expec(\resampprob{i}-\origprob{i})| &\leq \sqrt{Q_i}(C_1^2 + C_2) + C_3,\numberthis\label{eq:pi_1st_moment_C}\\
    \left|\sqrt{\expec\big((\resampprob{i}-\origprob{i})^2\big)}-\sqrt{Q_i}\right| &\leq \sqrt{Q_i}(C_1+C_2)+C_3 \numberthis\label{eq:pi_2nd_moment_C}
\end{align*}
where
\begin{align*}
    C_1 &= \sqrt{2dn}\exp\left(\|\origparam\|_2\xmax-\frac{c^2}{6d}\right),\\
    C_2 &= \frac{5.54c^2\xmax}{\sqrt{\eigmin}},\\
    C_3 &= 2d\exp\left(-\frac{c^2}{3d}\right).
\end{align*}

It is now time to choose an appropriate value for $c$ to control the values $C_1$, $C_2$, and $C_3$.  We set $c$ so that
\begin{align*}
    c^2 \defeq \frac{1}{3.1}\cdot\frac{\epsilon\sqrt{\eigmin}}{5.54\xmax}.\numberthis\label{eq:c-def}
\end{align*}
We previously imposed $c < \sqrt{\eigmin}/(40\xmax)$, which we will now verify.  Recall that from the statement of the theorem we have
\begin{align*}
    \epsilon \defeq 104\cdot\frac{d\xmax\big(\log(n\xmax/\xmin)+\xmax\|\origparam\|_2\big)}{\sqrt{\eigmin}}
\end{align*}
and by assumption $\epsilon < 1$.  This condition on $\epsilon$ implies $\sqrt{\eigmin}/\xmax > 100$, and therefore our choice of $c$ satisfies
\[c^2 < \frac{\sqrt{\eigmin}}{17\xmax} < \frac{\eigmin}{1700\xmax^2}.\]
We turn to the quantities $C_1$, $C_2$, and $C_3$.   By construction, $C_2 = \epsilon/3.1$.  To bound $C_1$ and $C_2$, we first note that
\begin{align*}
c^2 \geq 6d\xmax\left(\log(n\xmax/\xmin)+\xmax\|\origparam\|_2\right).
\end{align*}
It will be useful to have a bound for $\eigmin$ and $\eigmax$.  Since for any unit vector $\bm{u}\in\R^d$ we have
\[\bm{u}^\transpose \Hess \bm{u} = \sum_{j=1}^n \origprob{j}(1-\origprob{j})(\pt{j}^\transpose\bm{u})^2 \leq n \xmax^2,\]
it follows that 
$\eigmin \leq \eigmax \leq n \xmax^2$.  Hence, we have $\epsilon \geq 100d/\sqrt{n}$.  This permits the bound
\begin{align*}
6d\left(\|\origparam\|_2\xmax+\log\left(\frac{\sqrt{2dn}}{\epsilon/50}\right)\right) &\leq 6d\|\origparam\|_2\xmax + 6d\log (n) \leq c^2
\end{align*}
which implies $C_1 \leq \epsilon/50$.  To control $C_3$, we first bound
\begin{align*}
\sqrt{Q_i} &= \origprob{i}(1-\origprob{i})\normHinv{\pt{i}} \\
&= \frac{1}{1+\exp\left(\pt{i}^\transpose\origparam\right)}\cdot\frac{1}{1+\exp\left(-\pt{i}^\transpose\origparam\right)}\cdot\pt{i}^\transpose\invHess\pt{i} \\
&\geq \frac{1}{3+\exp\left(\xmax\|\origparam\|_2\right)}\cdot\frac{\xmin}{\sqrt{\eigmax}}.
\end{align*}
Using this bound and $\epsilon \geq 100d/\sqrt{n}$, we bound
\begin{align*}
3d\log\left(\frac{2d}{(\epsilon/50)\sqrt{Q_i}}\right) &\leq 3d\log\left(\frac{\sqrt{n\eigmax}\big(3+\exp\left(\xmax\|\origparam\|_2\right)\big)}{\xmin}\right) \\
&\leq 3d\left(\log(4)+\xmax\|\origparam\|_2 +\log\left(\frac{\sqrt{n\eigmax}}{\xmin}\right)\right) \\
&\leq 3d\left(\log(4)+\xmax\|\origparam\|_2 +\log\left(\frac{\sqrt{n\eigmax}}{\xmax}\right) +\log\left(\frac{\xmax}{\xmin}\right)\right)\\
&\leq 3d\left(\log(n)+\xmax\|\origparam\|_2 +\log\left(n\right) +\log\left(\frac{\xmax}{\xmin}\right)\right) \\
&\leq c^2.
\end{align*}
To derive the second line, we use that $\log(3+\exp(x))\leq\log(4)+x$ for $x\geq 0$, which follows by taking derivatives. In the fourth line, we use $n \geq 4$.  From this computation, we conclude that $C_3 \leq \epsilon\sqrt{Q_i}/50$.  The bounds from \cref{eq:pi_1st_moment_C,eq:pi_2nd_moment_C} then become
\begin{align*}
    |\expec(\resampprob{i}-\origprob{i})| &\leq (\epsilon/3.1+\epsilon/50 + (\epsilon/50)^2)\sqrt{Q_i},\\
    \left|\sqrt{\expec\big((\resampprob{i}-\origprob{i})^2\big)}-\sqrt{Q_i}\right| &\leq (\epsilon/3.1 + \epsilon/50 + \epsilon/50)\sqrt{Q_i}.
\end{align*}
Recall again that $\epsilon < 1$.  We conclude
\begin{align*}
    \expec(\resampprob{i}-\origprob{i})^2 &\leq \left(0.363\right)^2\epsilon Q_i,\\
    \left|\expec\big((\resampprob{i}-\origprob{i})^2\big)-Q_i\right| &\leq  \left(2\cdot0.363 + 0.363^2\right)\epsilon Q_i.
\end{align*}
Applying these to \cref{eq:var_decomp}, we obtain the bound
\begin{align*}
|\Var(\resampprob{i})-Q_i| \leq \epsilon Q_i.
\end{align*}
\end{proof}

\paragraph{Final remarks on \cref{thm:var_bd}}  The constant $104$ in the theorem is not optimal for the above argument, as throughout the proof above we used bounds that are not tight.  We have opted to present the argument in a way that yields a close-to-optimal constant without needlessly involving complicated terms.

\subsection{Proof of \texorpdfstring{\cref{cor:fixed-dist}}{Corollary 6.3.2}}\label{app:cor-proof}

Using the notation of \cref{sec:proof}, we now prove \cref{cor:fixed-dist}.

\corofixeddist*

\begin{proof}
    We use \cref{thm:var_bd}.  Since $\mathcal{P}$ has compact support on $\R^d \setminus \{0\}$, it is the case that $\xmax$ and $\xmin$ are almost surely bounded above and below by a constant that depends on $\mathcal{P}$.  We therefore only need to show that there exists constants $c_1$ and $c_2$ such that
    \begin{align}
        \|\origparam\|_2 &\leq c_1 \label{eq:bd-origparam}, \\
        \eigmin &\geq c_2 n\label{eq:bd-eigmin}.
    \end{align}
    We see that \cref{eq:bd-origparam} implies \cref{eq:bd-eigmin} with probability $1-\delta/2$.  To see this, we condition on \cref{eq:bd-origparam} holding. Then, we have
    \[\sum_{i=1}^n \origprob{i}(1-\origprob{i})\pt{i}\pt{i}^\transpose \succeq \frac{\exp(-2\xmax\|\origparam\|_2)}{4}\sum_{i=1}^n \pt{i}\pt{i}^\transpose \succeq \frac{\exp(-2C_1\xmax)}{4}\sum_{i=1}^n \pt{i}\pt{i}^\transpose,\]
    where $\mA \succeq \mB$ denotes that $\mA - \mB$ is positive semidefinite.  It therefore suffices to bound the eigenvalues of $\mX = \sum_{i=1}^n \pt{i}\pt{i}^\transpose$ below.  Let $\mu_{\min}$ and $\mu_{\max}$ be the smallest and largest eigenvalues of
    \[\expec_{(\pt{},y)\sim\mathcal{P}}(\pt{}\pt{}^\transpose),\]
    the matrix of second moments of $\pt{}$ when sampled from $\mathcal{P}$.  By assumption, $\mu_{\min} > 0$.
    From \citep[Exercise 4.7.3]{vershynin_2018}, there is a positive constant $c_3 > 1$ (depending on $\mathcal{P}$) such that for all $t \in \R$ the smallest eigenvalue of $\mX$ (denoted $m$) is bounded by
    \begin{equation}m \geq n\mu_{\min} - c_3 \left(d + t + \sqrt{n(d+t)}\right)\mu_{\max}
    \end{equation}
    with probability at least $1-2\exp(-t)$.  We can set 
    $t = \log(4/\delta)$.  Then, if
    \[n \geq \left(\frac{4c_3\mu_{\max}}{\mu_{\min}}\right)^2\left(\log\left(\frac{4}{\delta}\right)+d\right),\]
    we see that $m$ is bounded below by
    \begin{align*}
        m &\geq n\mu_{\min} - c_3 \left(d + \log(4/\delta) + \sqrt{n(d+\log(4/\delta))}\right)\mu_{\max} \\
        &\geq n\mu_{\min} - c_3 \left(2\sqrt{n\big(\log(4/\delta)+d\big)}\right)\mu_{\max} \\
        &\geq \frac{n\mu_{\min}}{2}
    \end{align*}
    with probability at least $1-\delta/2$.

    We now turn to verify \cref{eq:bd-origparam}.  Let $\mu_{\min}$ again be the smallest eigenvalue of the matrix of second moments of $\pt{}$ when sampled from $\mathcal{P}$.  By assumption, the set of probabilities $\prob(y = 1 \;|\; x = x')$ for $x'$ in this set takes values bounded away from $0$ and $1$.  Let $q$ be such that these probabilities are all in the interval $[q,1-q]$.  Now consider an arbitrary vector $\paramvec$ on the sphere of radius $R \defeq 12\xmax / (q\mu_{\min})$.  The value the logistic loss takes at $\paramvec$ is a random variable $L(\paramvec)$ depending on $\{(\pt{i},\origlabel{i})\}_{i=1}^n$ defined by
    \[L(\paramvec) = \sum_{i=1}^n \log(1+\exp(-\origlabel{i}\pt{i}^\transpose\paramvec)).\]
    We can bound 
    \[\expec (\paramvec^\transpose \pt{}\pt{}^\transpose\paramvec) \geq \mu_{\min} R^2.\]
    Since $|\pt{}^\transpose\paramvec|$ is a random variable that takes value in $[0,R \xmax]$, we know that
    \[\prob\left(|\pt{i}^\transpose \paramvec | \geq \frac{R\sqrt{\mu_{\min}}}{2}\right)\] 
    is at least $\sqrt{\mu_{\min}}/(2\xmax)$.  We bound the loss at $\paramvec$ below by
    \begin{align*}L(\paramvec) &\geq \sum_{i=1}^n\log\big(1+\exp(-\origlabel{i}\pt{i}^\transpose\paramvec)\big)\cdot\ind\big(|\pt{i}^\transpose\paramvec| \geq R\sqrt{\mu_{\min}}/2 \text{ and } \sign(\origlabel{i}\pt{i}^\transpose\paramvec) = -1\big) \\
    &\geq \sum_{i=1}^n R\sqrt{\mu_{\min}}/2\cdot\ind\big(|\pt{i}^\transpose\paramvec| \geq R\sqrt{\mu_{\min}}/2 \text{ and } \sign(\origlabel{i}\pt{i}^\transpose\paramvec) = -1\big).
    \end{align*}
    The random variables (for each $i$) \[\ind\big(|\pt{i}^\transpose\paramvec| \geq R\sqrt{\mu_{\min}}/2 \text{ and } \sign(\origlabel{i}\pt{i}^\transpose\paramvec) = -1\big)\]
    are each a Bernoulli random variable that takes value $1$ with probability at least
    \[\frac{q\sqrt{\mu_{\min}}}{2\xmax}.\]
    For $i\neq j$, these variables are independent.  The expected value of $L(\paramvec)$ is therefore at least
    \[\frac{Rq\mu_{\min}}{4\xmax} = 3n.\]
    By Hoeffding's inequality,
    \begin{align*}
    \prob\left( L(\paramvec) \leq 2n \right) \leq \exp\left(-\frac{2n^2}{R^2\mu_{\min}  n/4}\right) = \exp\left(-\frac{8n}{R^2\mu_{\min}}\right).
    \end{align*}
    Now consider another parameter vector $\paramvec'$ on the sphere of radius $R$ with $\|\paramvec' - \paramvec\| < \delta \defeq 1/\xmax$.  We see that
    \begin{align*}
        L(\paramvec') &= \sum_{i=1}^n \log\big(1+\exp(-\origlabel{i}\pt{i}^\transpose\paramvec')\big) \\
        &\geq \sum_{i=1}^n \log\bigg(\exp(-\xmax \delta)\big(1+\exp(-\origlabel{i}\pt{i}^\transpose\paramvec)\big)\bigg) \\
        &= L(\paramvec) - n \xmax \delta = L(\paramvec) - n.
    \end{align*}
    This implies that if $L(\paramvec) \geq 2n$, then $L(\paramvec') \geq n$ for all $\paramvec'$ such that $\|\paramvec'\| = R$ and $\|\paramvec-\paramvec'\| = 1/\xmax$.  We can cover the sphere with a set $\mathcal{N}$ consisting of $\left(1+2/\delta\right)^d$ points such that every point on the sphere is within distance $\delta$ from a point in $\mathcal{N}$~\citep[Corollary 4.2.13]{vershynin_2018}.  Fix such a set $\mathcal{N}$.  By a union bound, the probability that $L(\paramvec) \leq 2n$ for all $\paramvec \in \mathcal{N}$ is at most
    \[\left(1+2\xmax\right)^d \exp\left(-\frac{8n}{R^2\mu_{\min}}\right).\]
    This is at most $\delta/2$ if we take
    \[n \geq \frac{R^2\mu_{\min}}{8}\log\left(\frac{2(1+2\xmax)^d}{\delta}\right).\]
    However, the logistic loss is convex and attains the value $n\log 2 < n$ at $0$, so therefore in this case the minimum must occur within the ball of radius $R$.  Hence, $\|\origparam\| \leq R$ with probability at least $1-\delta/2$.  The probability \cref{eq:bd-origparam,eq:bd-eigmin} both hold is $(1-\delta/2)^2 \geq 1-\delta$.
\end{proof}

\section{Conclusion}\label{sec:conclusion}

In machine learning tasks, there are typically multiple valid models that can be adopted.  The choice of which model to implement can be considered arbitrary, which means that predictions which differ between the models are also arbitrary.  In this work, we introduce the concept of observational multiplicity, which is a form of model multiplicity that arises from randomness in the data observation process. Observational multiplicity arises through different mechanisms than previously studied forms of model multiplicity and is a phenomenon which produces different competing models, but these alternative models are still equally valid.  A particular form of observational multiplicity in the context of probabilistic classification, which occurs as a result of observing single draws of random labels.  We show this can produce variation in model predictions, which we call regret.  We provide a general-purpose algorithm for estimating regret and derive an explicit formula for regret in the case of well-specified logistic regression.  Our experiments demonstrate that, for logistic regression, random forest, and gradient boosting tree models, regret is not uniformly distributed in the population and that the range of possible regrets can be quite large even in large datasets.  Lastly, we demonstrate how measuring regret can aid in downstream tasks through active learning or selective prediction.

\acks{This material is based upon work supported by the National Science Foundation under Grant No. DMS-1928930 and by the Alfred P.\ Sloan Foundation under grant G-2021-16778, while the authors were in residence at the Simons Laufer Mathematical Sciences Institute (formerly MSRI) in Berkeley, California, during the Fall 2023 semester. EG and DN were partially supported by NSF DMS 2408912. EG was supported by NSF DGE 2034835.}

\clearpage
\bibliography{refs}

@article{apnea-dataset,
    author = {Berk Ustun and M. Brandon Westover and Cynthia Rudin  and Matt T. Bianchi },
    title = {Clinical Prediction Models for Sleep Apnea: The Importance of Medical History over Symptoms},
    journal = {Journal of Clinical Sleep Medicine},
    volume = {12},
    number = {02},
    pages = {161-168},
    year = {2016},
    doi = {10.5664/jcsm.5476},
    URL = {https://jcsm.aasm.org/doi/abs/10.5664/jcsm.5476},
    eprint = {https://jcsm.aasm.org/doi/pdf/10.5664/jcsm.5476}
}

@article{bach2010self,
  title={Self-concordant analysis for logistic regression},
  author={Bach, Francis},
  journal={Electronic Journal of Statistics},
  volume={4},
  pages={384--414},
  year={2010},
  doi={}
}

@article{bank-dataset,
title = {A data-driven approach to predict the success of bank telemarketing},
journal = {Decision Support Systems},
volume = {62},
pages = {22-31},
year = {2014},
noissn = {0167-9236},
doi = {10.1016/j.dss.2014.03.001},
url = {https://www.sciencedirect.com/science/article/pii/S016792361400061X},
author = {S{\'e}rgio Moro and Paulo Cortez and Paulo Rita}
}

@article{bousquet2002stability,
  author={Bousquet, Olivier and Elisseeff, Andr{\'e}},
  title={Stability and generalization},
  journal = {Journal of Machine Learning Research},
  year    = {2002},
  volume  = {2},
  number  = {Mar},
  pages   = {499--526},
  url     = {https://www.jmlr.org/papers/v2/bousquet02a.html}
}

@inproceedings{lei2022stability,
 author = {Lei, Yunwen and Jin, Rong and Ying, Yiming},
 booktitle = {Advances in Neural Information Processing Systems},
 editor = {S. Koyejo and S. Mohamed and A. Agarwal and D. Belgrave and K. Cho and A. Oh},
 pages = {38557--38570},
 publisher={Curran Associates, Inc.},
 title = {Stability and Generalization Analysis of Gradient Methods for Shallow Neural Networks},
 url = {https://proceedings.neurips.cc/paper_files/paper/2022/file/fb8fe6b79288f3d83696a5d276f4fc9d-Paper-Conference.pdf},
 volume = {35},
 year = {2022}
}

@misc{wolberg1995breast,
  author       = {Wolberg, William and Mangasarian, Olvi and Street, Nick and Street, W.},
  title        = {Breast Cancer {W}isconsin (Diagnostic)},
  year         = {1993},
  howpublished = {UCI Machine Learning Repository},
  doi          = {10.24432/C5DW2B}
}

@article{zhao2006model,
  title={On model selection consistency of Lasso},
  author={Zhao, Peng and Yu, Bin},
  journal={Journal of Machine Learning Research},
  volume={7},
  number={90},
  pages={2541--2563},
  year={2006},
  url     = {http://jmlr.org/papers/v7/zhao06a.html}
}

@book{bootstrapmethods,
place={Cambridge},
publisher={Cambridge University Press},
title={Bootstrap methods and their applicaiton},
year={1997},
author={Davison, A. C. and Hinkley, D. V.},
series={Cambridge Series on Statistical and Probabilistic Mathematics}
}

@book{agrestibook,
author={Alan Agresti},
year={2002},
publisher={John Wiley \& Sons, Inc.},
place={Hokoben, New Jersey},
title={Categorical Data Analysis}
}

@book{vershynin_2018, 
place={Cambridge}, 
series={Cambridge Series in Statistical and Probabilistic Mathematics}, 
title={High-Dimensional Probability: An Introduction with Applications in Data Science}, 
doi={10.1017/9781108231596}, 
url = {https://doi.org/10.1017/9781108231596}, 
publisher={Cambridge University Press}, 
author={Vershynin, Roman}, 
year={2018}, 
collection={Cambridge Series in Statistical and Probabilistic Mathematics}
}

@inproceedings{black2022model,
author = {Black, Emily and Raghavan, Manish and Barocas, Solon},
title = {Model Multiplicity: Opportunities, Concerns, and Solutions},
year = {2022},
isbn = {9781450393522},
publisher={Association for Computing Machinery},
address = {New York, NY, USA},
url = {https://doi.org/10.1145/3531146.3533149},
doi = {10.1145/3531146.3533149},
booktitle = {Proceedings of the 2022 ACM Conference on Fairness, Accountability, and Transparency},
pages = {850--863},
numpages = {14},
location = {Seoul, Republic of Korea},
series = {FAccT '22}
}

@InProceedings{li2023transformers,
  title = 	 {Transformers as Algorithms: Generalization and Stability in In-context Learning},
  author =       {Li, Yingcong and Ildiz, Muhammed Emrullah and Papailiopoulos, Dimitris and Oymak, Samet},
  booktitle = 	 {Proceedings of the 40th International Conference on Machine Learning},
  pages = 	 {19565--19594},
  year = 	 {2023},
  editor = 	 {Krause, Andreas and Brunskill, Emma and Cho, Kyunghyun and Engelhardt, Barbara and Sabato, Sivan and Scarlett, Jonathan},
  volume = 	 {202},
  series = 	 {Proceedings of Machine Learning Research},
  nomonth = 	 {23--29 July},
  publisher={Proceedings of Machine Learning Research},
  url = 	 {https://proceedings.mlr.press/v202/li23l.html}
}

@article{buhlmann2002bagging,
    author = {Bühlmann, Peter and Yu, Bin},
    title = {Analyzing bagging},
    journal = {The Anals of Statistics},
    volume = {30},
    number = {4},
    pages = {927--961},
    year = {2002},
    month = {8}
}

@article{kim2023blackbox,
    author = {Kim, Byol and Barber, Rina Foygel},
    title = {Black-box tests for algorithmic stability},
    journal = {Information and Inference: A Journal of the IMA},
    volume = {12},
    number = {4},
    pages = {2690-2719},
    year = {2023},
    month = {12},
    issn = {2049-8772},
    doi = {10.1093/imaiai/iaad039},
    url = {https://doi.org/10.1093/imaiai/iaad039},
    eprint = {https://academic.oup.com/imaiai/article-pdf/12/4/2690/54531021/iaad039.pdf},
}

@inproceedings{10.1145/3593013.3593998,
author = {Watson-Daniels, Jamelle and Barocas, Solon and Hofman, Jake M. and Chouldechova, Alexandra},
title = {Multi-Target Multiplicity: Flexibility and Fairness in Target Specification under Resource Constraints},
year = {2023},
isbn = {9798400701924},
publisher = {Association for Computing Machinery},
address = {New York, NY, USA},
url = {https://doi.org/10.1145/3593013.3593998},
doi = {10.1145/3593013.3593998},
booktitle = {Proceedings of the 2023 ACM Conference on Fairness, Accountability, and Transparency},
pages = {297–311},
numpages = {15},
location = {Chicago, IL, USA},
series = {FAccT '23}
}

@InProceedings{pmlr-v124-pawelczyk20a,
  title = 	 {On Counterfactual Explanations under Predictive Multiplicity},
  author =       {Pawelczyk, Martin and Broelemann, Klaus and Kasneci, Gjergji.},
  booktitle = 	 {Proceedings of the 36th Conference on Uncertainty in Artificial Intelligence (UAI)},
  pages = 	 {809--818},
  year = 	 {2020},
  editor = 	 {Peters, Jonas and Sontag, David},
  volume = 	 {124},
  series = 	 {Proceedings of Machine Learning Research},
  month = 	 {03--06 Aug},
  publisher =    {PMLR},
  url = 	 {https://proceedings.mlr.press/v124/pawelczyk20a.html}
}

@article{damour2020underspecification,
  author  = {Alexander D'Amour and Katherine Heller and Dan Moldovan and Ben Adlam and Babak Alipanahi and Alex Beutel and Christina Chen and Jonathan Deaton and Jacob Eisenstein and Matthew D. Hoffman and Farhad Hormozdiari and Neil Houlsby and Shaobo Hou and Ghassen Jerfel and Alan Karthikesalingam and Mario Lucic and Yian Ma and Cory McLean and Diana Mincu and Akinori Mitani and Andrea Montanari and Zachary Nado and Vivek Natarajan and Christopher Nielson and Thomas F. Osborne and Rajiv Raman and Kim Ramasamy and Rory Sayres and Jessica Schrouff and Martin Seneviratne and Shannon Sequeira and Harini Suresh and Victor Veitch and Max Vladymyrov and Xuezhi Wang and Kellie Webster and Steve Yadlowsky and Taedong Yun and Xiaohua Zhai and D. Sculley},
  title   = {Underspecification Presents Challenges for Credibility in Modern Machine Learning},
  journal = {Journal of Machine Learning Research},
  year    = {2022},
  volume  = {23},
  number  = {226},
  pages   = {1--61},
  url     = {http://jmlr.org/papers/v23/20-1335.html}
}

@article{nagaraj2025regretful,
  title={Regretful Decisions under Label Noise},
  author={Nagaraj, Sujay and Liu, Yang and Calmon, Flavio P and Ustun, Berk},
  journal={13th International Conference on Learning Representations},
  year={2025},
  url = {https://openreview.net/forum?id=7B9FCDoUzB}
}

@inproceedings{meyer2023dataset,
    author = {Meyer, Anna P. and Albarghouthi, Aws and D'Antoni, Loris},
    title = {The Dataset Multiplicity Problem: How Unreliable Data Impacts Predictions},
    year = {2023},
    isbn = {9798400701924},
    publisher={Association for Computing Machinery},
    address = {New York, NY, USA},
    url = {https://doi.org/10.1145/3593013.3593988},
    doi = {10.1145/3593013.3593988},
    booktitle = {Proceedings of the 2023 ACM Conference on Fairness, Accountability, and Transparency},
    pages = {193--204},
    numpages = {12},
    location = {Chicago, IL, USA},
    series = {FAccT '23}
}

@inproceedings{verma2019stability,
    author = {Verma, Saurabh and Zhang, Zhi-Li},
    title = {Stability and Generalization of Graph Convolutional Neural Networks},
    year = {2019},
    isbn = {9781450362016},
    publisher={Association for Computing Machinery},
    address = {New York, NY, USA},
    url = {https://doi.org/10.1145/3292500.3330956},
    doi = {10.1145/3292500.3330956},
    booktitle = {Proceedings of the 25th ACM SIGKDD International Conference on Knowledge Discovery \& Data Mining},
    pages = {1539--1548},
    numpages = {10},
    location = {Anchorage, AK, USA},
    series = {KDD '19}
}

@inproceedings{ganesh2025systemizing,
    author = {Ganesh, Prakhar and Taik, Afaf and Farnadi, Golnoosh},
    title = {Systemizing Multiplicity: The Curious Case of Arbitrariness in Machine Learning},
    booktitle = {Proceedings of the AAAI ACM Conference on AI, Ethics, and Society},
    year = {2025},
    month = {10},
    volume = {8},
    nonumber = {2},
    pages = {1032–-1048},
    doi = {https://doi.org/10.1609/aies.v8i2.36609}
}

@inproceedings{ICLR2025_94855277,
 author = {Du, Ally and Ngo, Dung Daniel and Wu, Steven},
 booktitle = {International Conference on Learning Representations},
 editor = {Y. Yue and A. Garg and N. Peng and F. Sha and R. Yu},
 pages = {59133--59164},
 title = {Reconciling Model Multiplicity for Downstream Decision Making},
 url = {https://proceedings.iclr.cc/paper_files/paper/2025/file/948552777302d3abf92415b1d7e9de70-Paper-Conference.pdf},
 volume = {2025},
 year = {2025}
}

@inproceedings{watson2023predictive,
    title={Predictive Multiplicity in Probabilistic Classification},
    volume={37},
    url={https://ojs.aaai.org/index.php/AAAI/article/view/26227},
    DOI={10.1609/aaai.v37i9.26227},
    abstractNote={Machine learning models are often used to inform real world risk assessment tasks: predicting consumer default risk, predicting whether a person suffers from a serious illness, or predicting a person’s risk to appear in court. Given multiple models that perform almost equally well for a prediction task, to what extent do predictions vary across these models? If predictions are relatively consistent for similar models, then the standard approach of choosing the model that optimizes a penalized loss suffices. But what if predictions vary significantly for similar models? In machine learning, this is referred to as predictive multiplicity i.e. the prevalence of conflicting predictions assigned by near-optimal competing models. In this paper, we present a framework for measuring predictive multiplicity in probabilistic classification (predicting the probability of a positive outcome). We introduce measures that capture the variation in risk estimates over the set of competing models, and develop optimization-based methods to compute these measures efficiently and reliably for convex empirical risk minimization problems. We demonstrate the incidence and prevalence of predictive multiplicity in real-world tasks. Further, we provide insight into how predictive multiplicity arises by analyzing the relationship between predictive multiplicity and data set characteristics (outliers, separability, and majority-minority structure). Our results emphasize the need to report predictive multiplicity more widely.},
    booktitle={Proceedings of the AAAI Conference on Artificial Intelligence},
    author={Watson-Daniels, Jamelle and Parkes, David C. and Ustun, Berk},
    year={2023},
    nomonth={June},
    pages={10306--10314},
    nonumber={9},
}

@misc{fannie_mae_dataset,
    author={{Fannie Mae}},
    title={{F}annie {M}ae Single-Family Loan Performance Data},
    url={https://capitalmarkets.fanniemae.com/credit-risk-transfer/single-family-credit-risk-transfer/fannie-mae-single-family-loan-performance-data},
    note={[Online; accessed: 2025-1-16]},
    year={2025}
}

@inproceedings{hsu2022rashomon,
 author = {Hsu, Hsiang and Calmon, Flavio},
 booktitle = {Advances in Neural Information Processing Systems},
 editor = {S. Koyejo and S. Mohamed and A. Agarwal and D. Belgrave and K. Cho and A. Oh},
 pages = {28988--29000},
 publisher={Curran Associates, Inc.},
 title = {Rashomon Capacity: A Metric for Predictive Multiplicity in Classification},
 url = {https://proceedings.neurips.cc/paper_files/paper/2022/file/ba4caa85ecdcafbf9102ab8ec384182d-Paper-Conference.pdf},
 volume = {35},
 year = {2022}
}

@InProceedings{marx2019predictive,
  title = 	 {Predictive Multiplicity in Classification},
  author =       {Marx, Charles and Calmon, Flavio and Ustun, Berk},
  booktitle = 	 {Proceedings of the 37th International Conference on Machine Learning},
  pages = 	 {6765--6774},
  year = 	 {2020},
  editor = 	 {Daumé, III, Hal and Singh, Aarti},
  volume = 	 {119},
  series = 	 {Proceedings of Machine Learning Research},
  nomonth = 	 {13--18 July},
  publisher={Proceedings of Machine Learning Research},
  url = 	 {https://proceedings.mlr.press/v119/marx20a.html}
}

@article{breiman2001statistical,
    author = {Leo Breiman},
    title = {Statistical Modeling: The Two Cultures},
    volume = {16},
    journal = {Statistical Science},
    number = {3},
    publisher={Institute of Mathematical Statistics},
    pages = {199--231},
    year = {2001},
    doi = {10.1214/ss/1009213726},
    URL = {https://doi.org/10.1214/ss/1009213726}
}

\end{document}